\documentclass[11pt]{article}

\usepackage{acl}
\usepackage{times}
\usepackage{latexsym}
\usepackage[T1]{fontenc}
\usepackage[utf8]{inputenc}
\usepackage{microtype}
\usepackage{inconsolata}
\usepackage{graphicx}
\usepackage{booktabs}
\usepackage{amsmath}
\usepackage{multirow}
\usepackage{CJKutf8}

\usepackage{tikz}
\usetikzlibrary{positioning}
\usepackage[table]{xcolor}

\definecolor{chggreen}{rgb}{0.0,0.50,0.0}

\definecolor{revred}{rgb}{0.80,0.00,0.00}

\definecolor{hevblue}{rgb}{0.00,0.25,0.75}

\usepackage{listings}
\lstdefinestyle{prompt}{
  basicstyle=\ttfamily\scriptsize,
  breaklines=true, breakatwhitespace=true, breakindent=0pt,
  columns=fullflexible, keepspaces=true,
  frame=single, framesep=5pt,
  rulecolor=\color{black!35}, backgroundcolor=\color{black!4},
  xleftmargin=0pt, xrightmargin=0pt,
  showstringspaces=false, aboveskip=4pt, belowskip=4pt,
}

\usepackage{cuted}
\usepackage{capt-of}
\usepackage[]{todonotes}

\newcommand{\Eng}[1]{English}
\newcommand{\Hye}[1]{Eastern Armenian}
\newcommand{\Bel}[1]{Belarusian}
\newcommand{\Zhs}[1]{Simplified Chinese}
\newcommand{\Zht}[1]{Traditional Chinese Taiwan}
\newcommand{\Ind}[1]{Indonesian}
\newcommand{\Kaz}[1]{Kazakh}
\newcommand{\Kor}[1]{Korean}
\newcommand{\Rus}[1]{Russian}
\newcommand{\Tha}[1]{Thai}
\newcommand{\Ukr}[1]{Ukrainian}
\newcommand{\Ces}[1]{Czech}
\newcommand{\Deu}[1]{German}
\newcommand{\Jpn}[1]{Japanese}
\newcommand{\Ekk}[1]{Estonian}
\newcommand{\Isl}[1]{Icelandic}

\newcommand{\EngHye}[1][]{\Eng{} $\rightarrow$ \Hye{}}
\newcommand{\EngBel}[1][]{\Eng{} $\rightarrow$ \Bel{}}
\newcommand{\EngZhs}[1][]{\Eng{} $\rightarrow$ \Zhs{}}
\newcommand{\EngZht}[1][]{\Eng{} $\rightarrow$ \Zht{}}
\newcommand{\EngInd}[1][]{\Eng{} $\rightarrow$ \Ind{}}
\newcommand{\EngKaz}[1][]{\Eng{} $\rightarrow$ \Kaz{}}
\newcommand{\EngKor}[1][]{\Eng{} $\rightarrow$ \Kor{}}
\newcommand{\EngRus}[1][]{\Eng{} $\rightarrow$ \Rus{}}
\newcommand{\EngTha}[1][]{\Eng{} $\rightarrow$ \Tha{}}
\newcommand{\EngUkr}[1][]{\Eng{} $\rightarrow$ \Ukr{}}
\newcommand{\CesDeu}[1][]{\Ces{} $\rightarrow$ \Deu{}}
\newcommand{\ZhsJpn}[1][]{\Zhs{} $\rightarrow$ \Jpn{}}
\newcommand{\EngEkk}[1][]{\Eng{} $\rightarrow$ \Ekk{}}
\newcommand{\EngDeu}[1][]{\Eng{} $\rightarrow$ \Deu{}}
\newcommand{\EngIsl}[1][]{\Eng{} $\rightarrow$ \Isl{}}
\newcommand{\EngJpn}[1][]{\Eng{} $\rightarrow$ \Jpn{}}

\newcommand{\eng}[1]{eng}
\newcommand{\hye}[1]{hye}
\newcommand{\bel}[1]{bel}
\newcommand{\zhs}[1]{zho-Hans}
\newcommand{\zht}[1]{zho-Hant}
\newcommand{\ind}[1]{ind}
\newcommand{\kaz}[1]{kaz}
\newcommand{\kor}[1]{kor}
\newcommand{\rus}[1]{rus}
\newcommand{\tha}[1]{tha}
\newcommand{\ukr}[1]{ukr}
\newcommand{\ces}[1]{ces}
\newcommand{\deu}[1]{deu}
\newcommand{\jpn}[1]{jpn}
\newcommand{\ekk}[1]{ekk}
\newcommand{\isl}[1]{isl}

\newcommand{\enghye}[1][]{\eng{} $\rightarrow$ \hye{}}
\newcommand{\engbel}[1][]{\eng{} $\rightarrow$ \bel{}}
\newcommand{\engzhs}[1][]{\eng{} $\rightarrow$ \zhs{}}
\newcommand{\engzht}[1][]{\eng{} $\rightarrow$ \zht{}}
\newcommand{\engind}[1][]{\eng{} $\rightarrow$ \ind{}}
\newcommand{\engkaz}[1][]{\eng{} $\rightarrow$ \kaz{}}
\newcommand{\engkor}[1][]{\eng{} $\rightarrow$ \kor{}}
\newcommand{\engrus}[1][]{\eng{} $\rightarrow$ \rus{}}
\newcommand{\engtha}[1][]{\eng{} $\rightarrow$ \tha{}}
\newcommand{\engukr}[1][]{\eng{} $\rightarrow$ \ukr{}}
\newcommand{\cesdeu}[1][]{\ces{} $\rightarrow$ \deu{}}
\newcommand{\zhsjpn}[1][]{\zhs{} $\rightarrow$ \jpn{}}
\newcommand{\engekk}[1][]{\eng{} $\rightarrow$ \ekk{}}
\newcommand{\engdeu}[1][]{\eng{} $\rightarrow$ \deu{}}
\newcommand{\engisl}[1][]{\eng{} $\rightarrow$ \isl{}}
\newcommand{\engjpn}[1][]{\eng{} $\rightarrow$ \jpn{}}

\title{In the Blind: Building Pseudo-References for MT Evaluation}

\author{Diptesh Kanojia$^{(1)}$,
  Chi-kiu Lo  \begin{CJK*}{UTF8}{bsmi}羅致翹\end{CJK*}$^{(2)}$,
  Archchana Sindhujan$^{(1)}$\\
  \bf{Samuel Larkin$^{(2)}$, 
  Greg Hanneman$^{(3)}$,
  Alon Lavie$^{(4)}$}\\
  $^{(1)}$University of Surrey
  $^{(2)}$National Research Council Canada\\
  $^{(3)}$Unaffiliated
  $^{(4)}$Carnegie Mellon University\\
  \tt{d.kanojia@surrey.ac.uk}
  }

\begin{document}
\maketitle

\begin{abstract}

The WMT26 General MT task evaluates systems on 10 language pairs that have no human references (neither translated from scratch nor post-edited from MT output by humans). We describe how we built the pseudo-references for these pairs and six other language pairs (in which some forms of human references are available): seven models translate the 3,277 official documents under up to five prompt conditions, giving a total of 26 system--prompt combinations; then three reference-free quality estimation (QE) models score every candidate; and a per-document selector picks one translation, which GPT-5.5 post-edits where needed. Working without references exposed a failure mode of QE-guided selection: the metrics rank fluent output in the wrong language above correct translations. Adding a confidence-scaled language identification penalty to the score fusion drives the wrong-language count to zero, and the resulting selector still scores better on MetricX than the rank-fusion baseline it replaces. Since no references were available for these pairs while we were building them, we calibrate every selection decision on last year's WMT25 human judgments. The human evaluation, released after construction, shows the cost of getting selection wrong: our references stand with the strongest participating systems when the selector kept a frontier-model candidate, and fall up to 17 ESA points below them when it did not. We release the selection method and the provenance of every reference.\footnote{\url{https://github.com/surrey-nlp/PseudoRef}}

\end{abstract}

\section{Introduction}
\label{sec:intro}
 
The rise in performance of automatic machine translation (MT) reference-less quality evaluation (commonly referred to as quality estimation, QE) metrics has put the necessity of references in MT evaluation and development in debate. Recent work in the WMT25 unified evaluation and quality estimation task \citep{lavie-etal-2025-findings} finds that reference-based metrics, including traditional string-based, embedding-based, and LLM-based, achieve top-ranking results in correlation with human judgments on translation quality. Reference-based metrics guide system development and shared-task rankings, and reference quality directly affects the conclusions drawn from evaluation results \citep{freitag-etal-2020-bleu}. However, producing human references is costly, time-consuming, and difficult to scale as evaluation expands to more language pairs. A possible solution to the unavailability of human references is using MT output of high quality as reference in evaluation which we called pseudo-references. Pseudo-references offer a practical alternative \citep{albrecht-hwa-2007-regression}.

\begin{figure}[t]
  \centering
  \begin{tikzpicture}[
      node distance=2.6mm,
      box/.style={draw, rounded corners=1.5pt, align=left, inner sep=3pt,
                  text width=0.97\columnwidth, font=\scriptsize},
      arr/.style={-latex, semithick}]
    \node[box] (s1) {\textbf{1. Generate: } 7 models $\times$ prompt conditions $\rightarrow$ 26 cells, 80{,}063 candidate documents};
    \node[box, below=of s1] (s2) {\textbf{2. Score:} MetricX-24, CometKiwi-XL, xCOMET-XL, reference-free, unit level (+ MQM spans)};
    \node[box, below=of s2] (s3) {\textbf{3.  Select:} structure gate $\rightarrow$ per-document $z$-fusion led by MetricX $\rightarrow$ language penalty $\rightarrow$ span tiebreak};
    \node[box, below=of s3] (s4) {\textbf{4.  Repair:} Constrained LLM post-editing, applied where selection is weak};
    \node[box, below=of s4] (s5) {\textbf{5.  Verify:}  Alignment checker; provenance record per document};
    \draw[arr] (s1) -- (s2);
    \draw[arr] (s2) -- (s3);
    \draw[arr] (s3) -- (s4);
    \draw[arr] (s4) -- (s5);
  \end{tikzpicture}
  \caption{The pseudo-reference pipeline. }
  \label{fig:pipeline}
\end{figure}
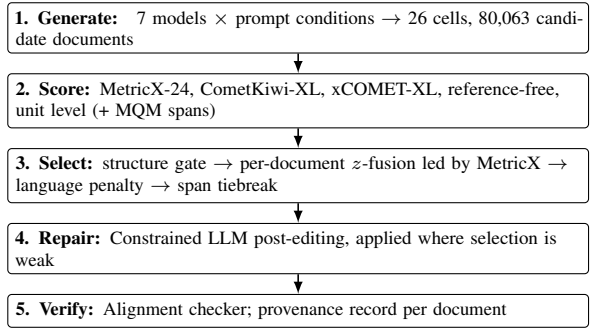

The obvious approach is to generate several candidate translations and keep the one a QE metric scores the highest. That assumes the selection signal is reliable. Neural metrics have systematic weaknesses that optimisation can exploit~\citep{amrhein-sennrich-2022-identifying}, their agreement with humans varies across quality levels, and the systems producing the candidates may have been tuned on the same metric families used to select among them. Any bias in selection is then baked into the references and propagates to every evaluation that uses them. We therefore treat reference construction and metric auditing as one task. Figure~\ref{fig:pipeline} shows the pipeline. Seven models generate candidates under several prompt conditions, three reference-free metrics score them, a per-document selector keeps one, and an LLM post-editor repairs it where the selection is weak. Section~\ref{sec:pipeline} gives each stage. We entered our selections into the General MT task as a participating team, where the human evaluation scored our primary submission as \texttt{PseudoRef} (Section~\ref{sec:refeval}, Appendix~\ref{app:humaneval}).

 
Our main finding is a failure that none of the three metrics detects. A language identifier run over the selector output found 154 of the 3{,}277 selected documents in the wrong target language, mostly fluent Russian or Ukrainian instead of Kazakh, Belarusian or Armenian, several of them scored best in their pools. \citet{amrhein-sennrich-2022-identifying} induced this behaviour deliberately with minimum Bayes risk (MBR) decoding; it arises on its own once a multilingual candidate pool meets a reference-free metric. A confidence-scaled language-identification penalty in $z$-score fusion removes all 154, improving mean MetricX.



Which metric to optimise is an empirical question, so we answered it on WMT25, re-scoring 25{,}824 human-judged candidates over the seven pairs that overlap ours. MetricX tracks human system rankings best; CometKiwi and xCOMET collapse among top candidates, which is where selection operates. These results, not the WMT26 QE scores, set our fusion weights and demote error spans to a tiebreak (Section~\ref{sec:calibration}). Table \ref{tab:pairs} shows these 16 pairs. Ten of our language pairs do not have any form of human-involved references; the remaining six language pairs are chosen to support analyses of the effects that reference quality has on automatic evaluation, as conflicting evidence is presented in previous work \citep{freitag-etal-2020-bleu,mathur-etal-2020-results,freitag-etal-2023-results}. The chosen six language pairs cover different source languages, resource levels and target orthographies. 

\begin{table}[t]
  \centering
  \small
  \setlength{\tabcolsep}{3pt}
  \begin{tabular}{@{}lr@{\hspace{2em}}|lr@{}}
    \toprule
    \cmidrule(r){1-2}\cmidrule(l){3-4}
    \textbf{Pair} & \textbf{Docs} & \textbf{Domain} & \textbf{Docs} \\
    \midrule
    \eng{}$\rightarrow$\bel{}, \deu{}, \ekk, \hye{}, \ind{}, \isl{}, & \multirow{3}{*}{198} & spoken & 2,090 \\
    \quad  \jpn{}, \kaz{}, \kor{}, \rus{}, \tha{}, \ukr{},      &                          & social & 737 \\
    \quad   \zhs{}, \zht{}&                          & news & 259 \\ 
    \ces{}$\rightarrow$\deu{} & 315 & software & 98 \\ 
    \zhs{}$\rightarrow$\jpn{} & 190 & edu & 63 \\
                      &     & factcheck & 30 \\
    \bottomrule
  \end{tabular}
  \caption{Documents by language pair and by domain.}
  \label{tab:pairs}
\end{table}

\section{Task Setting and Data}
\label{sec:task}


\paragraph{The WMT26 GenMT blindset.} The released blindset instances from WMT26~\citep{kocmi-etal-2026-findings} each contained a source document paired with a target language and an
instruction that is part of the graded task. The official core for pseudo-reference construction is 3,277 documents across the 16 pairs listed in Table~\ref{tab:pairs}: 198 documents for each of the 14 English-source pairs, 315 for \CesDeu{}, and 190 for \ZhsJpn{}. Six domains are represented: spoken transcripts, social media threads, news, software strings, education, and fact-checking articles. A 2,706-document subset is multimodal: spoken documents are ASR transcripts of videos, social documents come with screenshots, and annotators judge translations against the media rather than the text. Documents are structured, with HTML in social and news and JSON in software, and the official checker requires the reference to preserve markup element counts exactly.

\paragraph{Language identification.} We label every candidate with py3langid~\citep{lui-baldwin-2012-langid}: deterministic, offline, 97 languages, and it returns a calibrated confidence that Eq.~\ref{eq:penalty} uses directly. fastText~\cite{bojanowski2017enriching} and GlotLID~\cite{kargaran-etal-2023-glotlid} are reasonable alternatives, GlotLID with wider low-resource coverage; we did not benchmark the three against each other, which is a limitation of the gate rather than of the selector. Two of our targets are known confusion cases: Indonesian against Malay, where ours mislabels correct output (Section~\ref{sec:gaming}), and Japanese against Chinese in kanji-heavy text.


\section{The Pseudo-Reference Pipeline}
\label{sec:pipeline}

\begin{table}[t]
  \centering
  \small
  \renewcommand{\arraystretch}{1.08}
  \setlength{\tabcolsep}{5pt}
  \begin{tabular}{@{}llr@{}}
    \toprule
    \textbf{Model} & \textbf{Prompt conditions} & \textbf{Cells} \\
    \midrule

    \rowcolor{gray!15}
    \multicolumn{3}{@{}l}{\textbf{Closed }} \\

    GPT-5.5        & P1--P3, P4$\times 2$ & 5 \\
    Gemini-3.1-Pro & P1--P3, P4$\times 2$ & 5 \\

    \addlinespace[2pt]
    \rowcolor{gray!15}
    \multicolumn{3}{@{}l}{\textbf{Open-weight }} \\

    Aya-Expanse-32B    & P1--P3, P4$\times 2$ & 5 \\
    Tower-Plus-9B      & P1--P3, P4$\times 2$ & 5 \\
    EuroLLM-22B        & P1--P3, P4$\times 1$ & 4 \\
    TranslateGemma-27B & Native template       & 1 \\
    HY-MT1.5-7B        & Native template       & 1 \\

    \midrule
    \textbf{Total} & & \textbf{26} \\
    \bottomrule
  \end{tabular}

  \caption{Candidate-generation: A \emph{cell} denotes one
  model--prompt configuration.
  }
  \label{tab:panel}
\end{table}

\subsection{Candidate generation}
\label{sec:s1}

Seven models generate candidates (Table~\ref{tab:panel}) - five are open-weight: Aya-Expanse-32B \citep{dang-etal-2024-aya}, EuroLLM-22B \citep{martins-etal-2024-eurollm}, Tower-Plus-9B \citep{alves-etal-2024-tower, rei-etal-2025-towerplus}, and two dedicated translation models run under their native templates, TranslateGemma-27B \citep{finkelstein-etal-2026-translategemma} and HY-MT1.5-7B \citep{zheng-etal-2025-hymt}. Two are closed: GPT-5.5\footnote{Model version: 2026-04-24; accessed through the National Research Council of Canada’s Azure cloud platform} \citep{openai-2025-gpt55} and Gemini-3.1-pro \citep{google-2026-gemini3}. 
%

The instruction-following models run under five prompt (P) conditions. P1 is the 
verbatim of the document-level translation prompt provided by the WMT26 GenMT organizers; since instruction adherence is graded and shown to annotators, our additions never override it. P2 wraps this instruction in a source analysis stage before translating. P3 adds five-shot examples over P2. P4 adds the multimodal context to P3 as a text caption of the video or screenshot, produced by Qwen3-Omni \citep{xu-etal-2025-qwen3omni} or Gemini-3.5-Flash. P4 runs only on the 2,706-document multimodal subset. Prompts and configurations are present in Appendix~\ref{app:gen-prompts}. The caption is an English description of the media generated from the video or screenshot alone, prepended to the source document as background context, the translation model never sees the media itself. We run the two caption sources as separate cells to test whether the gain depends on the captioner rather than on captioning as such. The native-template models take a single pass each. We refer to each (model, prompt condition) pairing as a \emph{cell}: one complete generation run over the document set, contributing one candidate translation per document, totalling 26 scored cells as shown in Table~\ref{tab:panel}.


Not every cell produces a usable translation everywhere. The failures are model capability limits, concentrated on non-Latin targets, and Appendix~\ref{app:panel} gives them per cell. We left failed outputs in the pool, since discarding them is the selector's job.


\subsection{Reference-free QE}
\label{sec:s2}

Three QE models score every candidate: MetricX-24 (regression on an error scale, 0 to 25, lower is better), CometKiwi-XL (0 to 1, higher is better), and xCOMET-XL (0 to 1, higher is better, with MQM-style error spans \citep{lommel-etal-2014-mqm} as a by-product, each graded minor, major, or critical). Scoring is unit-level: spoken documents are scored per-utterance (7.8 units per document on average), software per element (17.4), and other domains per paragraph or block, with HTML stripped for scoring only. Document scores are unit means. Each metric scored 80{,}063 candidate documents, of which 79{,}971 are scorable; the rest are too broken to segment. Over the full candidate pool the three metrics agree closely with one another (pairwise Pearson 0.79 to 0.87), but that agreement comes from distinguishing usable translations from broken ones rather than from ranking the usable ones.

\subsection{Baseline selectors and structure gating}
\label{sec:s3}

The selector picks one candidate per document. We built two baseline selectors. \emph{Rank-fusion} ranks candidates under each metric separately and keeps the best mean rank, a Borda count over MetricX-24 and CometKiwi-XL; xCOMET contributes error spans but no rank. Using ranks rather than raw scores makes it insensitive to the fact that the metrics run on different scales and in different directions. \emph{Cross-reference} re-ranks that selector's top three by cross-system MBR: each is scored by reference-based COMET \citep{rei-etal-2022-comet} against the other two as pseudo-references, and the one with the highest mean score is kept.

Both are structure-gated where the official checker requires structured documents to preserve the source HTML paragraph or JSON element count exactly, and the QE metrics ignore markup. On the 1,191 structured documents, the gate admits 62.6\% of candidates and leaves every document at least two, so gating fixes alignment without ever having to ship a structurally invalid candidate; on 39 documents none of cross-reference's three was valid and the rank-fusion selection stands.
The two selectors agree on only 46\% of the documents they ship, and since one chooses within the other's top three, that is a first sign of how unstable selection among near-tied candidates is.  Both selectors are superseded by the language-aware selector of Section~\ref{sec:gaming}, for reasons the next section sets out.

\subsection{Automatic post-editing}
\label{sec:s4}

GPT-5.5 post-edits the rank-fusion selection, given the source, the top three candidates and their QE scores. Two earlier prompts, one showing scores alone (\textit{APE-scores}) and one adding xCOMET spans (\textit{APE-spans}), rewrote 99.5 and 99.9 percent of documents at median similarity of 0.59 and 0.56, respectively, which defeats the premise that the selection is mostly correct.
Both lost content anchors, 0.863 in the input against 0.839 and 0.834 out, a little worse with spans: a flagged number invites the model to correct what was already right. Worse, neither reliably repaired the wrong-language inputs it was given: of the 154 the langid audit finds in the rank-fusion selection, 21 survive APE-scores and 24 survive APE-spans. \textit{APE-tight} therefore pins the target language, names the confusable neighbour for each pair, reframes the QE scores as noisy pointers, and imposes a minimal-edit contract. It is the only run to preserve content anchors better than its own input (0.911). Appendix~\ref{app:ape-behaviour} compares the three runs.

\subsection{Verification and the final reference set}
\label{sec:s5}

Every output file is checked with the alignment script\footnote{\texttt{genmt\_check\_alignment.py} from
\url{https://github.com/wmt-conference/wmt-collect-translations}} provided by the WMT26 GenMT shared task, and all selections reported check clean: 1,187 structured documents verified, 0 misaligned (the 2,090 plain-text documents are outside the tool's scope). 
Four Japanese-target documents were generated under both \engjpn{} and \zhsjpn{}. The organisers list each \texttt{doc\_id} once, so we kept the single Chinese-source copy; these four are the difference between the 1,191 structured documents of Section~\ref{sec:s3} and the 1,187 checked here. We entered seven systems into the General MT task. The language-aware selection of Section~\ref{sec:selection} is the primary, and the only one the human evaluation scored; the rank-fusion and cross-reference selections, the best single open and closed systems, and two pooled variants were contrastive.

\section{Wrong-Language Identification}
\label{sec:gaming}

We audited the rank-fusion winners with py3langid \citep{lui-baldwin-2012-langid}, comparing the predicted language of each selected document with the expected target.
Table~\ref{tab:wronglang} gives the result: 154 of 3,277 winners are predicted to be in the wrong language, and they concentrate on the low-resource targets, each replaced by a higher-resource language of the same region or script.
Of 198 \engkaz{} winners 54 are in the wrong language, 51 of them Russian; of 198 \engbel{} winners 63 are, 50 of them Ukrainian and 13 Russian; and 14 of 198 \enghye{} winners, 13 of them Russian. The substitute need not be a relative: Kazakh is Turkic, and Armenian is not even written in Cyrillic. What these targets share is a neighbouring language with far more training data behind it. By domain, 127 of the 154 are spoken documents. The \engind{} counts are an Indonesian/Malay artefact: the text is Indonesian, so the genuine rank-fusion count is 145. We quote 154 for comparability with the same audit run over the other selectors. The artefact has a cost, since our penalty then demotes correct Indonesian candidates. A language gate is only as good as its identifier.



Neither of the two straightforward alternatives fully resolves the issue. Cross-reference selection reduces the number of wrong-language outputs from 154 to 87, but does not eliminate them. When multiple candidates are produced in the same incorrect language, consensus-based scoring may treat their agreement as supporting evidence and rank them more highly. Automatic post-editing is more effective: the tightened prompt reduces the number of wrong-language outputs to 19, 18 of which are attributable to the \engind{} language-identification error. However, this approach addresses the problem only after an incorrect candidate has already been selected and rewrites more than 98\% of documents to correct an error affecting fewer than 5\%. Because the failure arises during candidate selection, we address it directly at the selection stage.

\begin{table}[t]
  \centering
  \resizebox{\columnwidth}{!}{%
  \begin{tabular}{@{}lrrrrr@{}}
    \toprule
    Pair & N & \ RF & \ XR & APE5 & C6 \\
    \midrule
    \engkaz{} & 198 & 54 & 36 & 0 & 0 \\
    \engbel{} & 198 & 63 & 29 & 0 & 0 \\
    \enghye{} & 198 & 14 & 1 & 0 & 0 \\
    \zhsjpn{} & 190 & 4 & 1 & 0 & 0 \\
    \engekk{} & 198 & 3 & 2 & 0 & 0 \\
    \engtha{} & 198 & 3 & 1 & 0 & 0 \\
    \cesdeu{} & 315 & 2 & 2 & 0 & 0 \\
    \engzhs{}, \zht & 396 & 2 & 2 & 1 & 0 \\
    \engind{}$^{\dagger}$ & 198 & 9 & 13 & 18 & 0 \\
    other 6 pairs & 1,188 & 0 & 0 & 0 & 0 \\
    \midrule
    Total & 3,277 & 154 & 87 & 19 & 0 \\
    \bottomrule
  \end{tabular}}
  \caption{Wrong-language documents per selector: langid prediction differs from the expected target. Rank-Fusion, Cross-reference, APE5 post-edited (APE-tight), C6 the language-aware selector of Section~\ref{sec:selection}. $^{\dagger}$\engind{} counts are an ind/msa identifier artefact. 
  }

  \label{tab:wronglang}
\end{table}

\subsection{Language-aware selection}
\label{sec:selection}

The selector we submitted as our primary basis, called C6, works per document. Let $C$ be the structure-valid candidate pool and, for candidate $c$, let $x_m(c)$ be its score under metric $m$, oriented so higher is better (MetricX negated). 
Scores are standardised within the pool, $z_m(c) = \frac{x_m(c) - \mu_m}{\sigma_m}$, and fused:
\begin{equation}
  F(c) = \sum_{m} w_m\, z_m(c),
  \label{eq:fusion}
\end{equation}
with weights set from the calibration in Section~\ref{sec:calibration}. C6 uses MetricX only ($w = 1, 0, 0$). A soft language penalty is subtracted: If the identifier labels $c$ with language $\ell(c)$ at confidence $p(c)$ and $\ell(c)$ differs from the expected target, then
\begin{equation}
  F(c) \leftarrow F(c) - \lambda\, p(c), \qquad \lambda = 3.
  \label{eq:penalty}
\end{equation}

Candidates whose fused scores fall within $\varepsilon = 0.10$ of the best score are treated as tied. Ties are resolved in order by: (1) fewest xCOMET critical-error spans;
(2) highest cross-reference score, that is, the mean reference-based COMET score of the candidate against every other candidate in the pool, as in Section~\ref{sec:s3}; and (3) best balance of content-anchor preservation and
fluency (Appendix~\ref{app:stats}). We then apply a conditional repair step in two cases. First, if the selected candidate is in the wrong language, it is replaced with the \textit{APE-tight} output only when the post-edited text is in the target language and satisfies the structural-validity constraints. Second, if the selected candidate falls within the worst 5\% of MetricX scores for its language pair, it is replaced only when the post-edited output achieves an equal or better MetricX score. If no structure-valid candidate is available, the procedure falls back to the original rank-fusion selection. In the final run, the repair step replaced 29 low-estimated-quality candidates and no wrong-language candidates, because the language-aware penalty had already excluded wrong-language outputs during selection (Example in Appendix~\ref{app:example}).

\begin{table}[t]
  \centering
  \resizebox{\columnwidth}{!}{%
  \begin{tabular}{@{}lccccc@{}}
    \toprule
    Selector & MX$\downarrow$ & CK$\uparrow$ & xC$\uparrow$ & WL$\downarrow$ & Crit$\downarrow$ \\
    \midrule
    Rank-fusion & 4.068 & \textbf{0.670} & 0.697 & 154 & 5,516 \\
    APE5 (post-edit all) & 4.515 & 0.644 & 0.678 & 19 & 5,595 \\
    \midrule
    C6: MX + span tiebreak & 4.005 & 0.654 & 0.691 & \textbf{0} & 4,129 \\
    C2: MX only & \textbf{4.004} & 0.655 & 0.690 & \textbf{0} & 5,196 \\
    C5: hard gate & 4.067 & 0.663 & 0.698 & \textbf{0} & 4,450 \\
    C: 0.70/0.15/0.15 & 4.071 & 0.662 & 0.698 & \textbf{0} & 4,431 \\
    C3: 0.68/0.10/0.23 & 4.098 & 0.662 & 0.703 & \textbf{0} & 3,541 \\
    C4: equal weights & 4.236 & 0.669 & \textbf{0.708} & \textbf{0} & \textbf{3,415} \\
    \bottomrule
  \end{tabular}}
  \caption{
  Selector variants over all 3,277 documents.
  No variant is best on every measure; C6 is chosen for the MetricX near-optimum at zero wrong-language.
  MX: MetricX mean (lower better); CK: CometKiwi; xC: xCOMET; WL: wrong-language documents; Crit: xCOMET critical spans.
  Weight triples are MetricX/CometKiwi/xCOMET in Eq.~\ref{eq:fusion}.
  }
  \label{tab:variants}
\end{table}

Table~\ref{tab:variants} compares the weighting schemes. The C-family shares the structure gate, the language penalty, and the repair step of C6, differing only in the fusion weights $w$ of Eq.~\ref{eq:fusion} and in how the language constraint is enforced. C weights MetricX, CometKiwi and xCOMET 0.70/0.15/0.15; C3 uses 0.68/0.10/0.23, set in proportion to how well each metric agrees with human judgement (Section~\ref{sec:calibration}); C4 weights the three equally. C2 and C6 use MetricX alone, C6 adding the critical-span tiebreak. C5 keeps C's weights but replaces the soft penalty with a hard gate that drops wrong-language candidates from the pool outright.

C6 reaches a mean MetricX of 4.005, marginally behind the best variant (C2, MetricX only, at 4.004) and well ahead of the unconstrained selector's 4.068. C6 differs from C2 only in breaking near-ties on critical spans, which cuts them from 5{,}196 to 4{,}129 at a cost of 0.001 MetricX. C3 and C4 cut spans further but only by upweighting xCOMET and CometKiwi, which Section~\ref{sec:calibration} shows to be unreliable selection objectives. 

\begin{table}[t]
  \centering
  \small
  \begin{tabular}{@{}lrrrr@{}}
    \toprule
    Pair & ESA & MX$\downarrow$ & CK$\uparrow$ & xC$\uparrow$ \\
    \midrule
    \cesdeu{} & 98.7 & 3.98 & 0.550 & 0.446 \\
    \engekk{} & 88.9 & 8.26 & 0.697 & 0.564 \\
    \engisl{} & 83.4 & 7.57 & 0.647 & 0.520 \\
    \engjpn{} & 98.4 & 6.11 & 0.681 & 0.513 \\
    \engrus{} & 95.1 & 5.98 & 0.637 & 0.582 \\
    \engukr{} & 95.6 & 6.36 & 0.626 & 0.540 \\
    \engzhs{} & 98.9 & 5.03 & 0.650 & 0.463 \\
    \midrule
    all & 94.2 & 6.13 & 0.638 & 0.517 \\
    \bottomrule
  \end{tabular}
  \caption{What human-quality translation scores: mean human ESA and our three metrics on the human-best WMT25 candidates (1,409 paragraphs, 7 pairs).}
  \label{tab:humanbest}
\end{table}

The ablation shows what each component costs, measured against C6. Setting $\lambda = 0$ in Eq.~\ref{eq:penalty}, which disables the language penalty, returns 162 documents with the wrong language for a marginal gain on MetricX (0.05). We accept that trade because the swap costs almost nothing on MetricX: wrong-language winners beat their best correct-language rival by only a small margin, and C6 with the penalty remains 0.063 better than rank fusion. The structure gate is not optional, since misaligned documents are rejected outright.

C6 continues to select outputs from a diverse set of systems. TranslateGemma contributes 809 selected documents, followed by GPT-5.5 with 764, Gemini with 673, HY-MT with 559, Tower-Plus with 189, Aya with 178, and EuroLLM with 76. A further 29 documents use post-edited outputs. Overall, 55\% of the selected references come from open models, 44\% from closed models, and 1\% from post-editing. The selection entropy is 2.54 bits out of a maximum of 3.00 bits, showing that no single system dominates the final reference set.

\section{External Calibration on WMT25}
\label{sec:calibration}

This section answers one question: which metric should the selector optimise?
The answer is MetricX alone, with xCOMET spans used only to break near-ties,
because MetricX is the only one of the three that tracks human system rankings.
The evidence follows. We calibrate on the WMT25 GenMT human evaluation \citep{kocmi-etal-2025-findings}, which provides human references and human judgments (ESA, 0 to 100, with error spans; \citealp{kocmi-etal-2024-error}) for seven pairs that overlap ours: \cesdeu{}, \engekk{}, \engisl{}, \engjpn{}, \engrus{}, \engukr{}, and \engzhs{}. Because the two test sets differ, these numbers support relative comparisons only.

\paragraph{Calibrating the metric scales.}
We identified the human-best candidate for each paragraph in the WMT25 evaluation set and scored the resulting 1{,}409 translations with our metrics (Table~\ref{tab:humanbest}).
Their human ESA scores span 83 to 99, so these are translations human judges rated as good or near-perfect.
On our metrics, the same text scores MetricX 4.0 to 8.3 and xCOMET 0.45 to 0.58.
Our WMT26 selections, at a mean MetricX of 4.0, and their post-edited counterparts at 4.5, both fall inside the band that human-preferred WMT25 translation occupies, which is also the band in which these metrics discriminate least.

\begin{table}[t]
  \centering
  \small
  \resizebox{\columnwidth}{!}{%
  \begin{tabular}{@{}lcccc@{}}
    \toprule
    Metric & $r$ top & $\tau$ sys & Pairwise \% & Fitted $w$ \\
    \midrule
    MetricX-24 & \textbf{0.495} & \textbf{0.29} & \textbf{58.3} & \textbf{0.51} \\
    CometKiwi-XL & 0.071 & 0.17 & 56.6 & 0.42 \\
    xCOMET-XL & 0.166 & 0.12 & 55.9 & $-$0.07 \\
    3-metric fusion & -- & 0.26 & -- & -- \\
    \bottomrule
  \end{tabular}}
\caption{Agreement between each QE metric and WMT25 human judgment, computed on WMT25 data. $r$ top: Pearson against ESA on human-best candidates
($n{=}1{,}409$); $\tau$ sys: mean per-pair Kendall $\tau$ of system rankings
(37 systems, 7 pairs, $n{=}25{,}824$); Pairwise: segment-level pairwise
accuracy; Fitted $w$: normalised OLS weight.
}
  \label{tab:metricshuman}
\end{table}

\subsection{Metric agreement with human judgment}

Selection operates among candidates that are already of relatively high quality. We therefore evaluate metric--human agreement in two settings: among the human-best candidate for each paragraph and across the full set of human-scored WMT25 candidates (Table~\ref{tab:metricshuman}; per-pair results are provided in Appendix~\ref{app:wmt25}). Among the human-best candidates, MetricX correlates with ESA at ($r$=0.495), whereas CometKiwi ($r$=0.071) and xCOMET ($r$=0.166) show substantially weaker agreement. Although restricting the analysis to high-quality candidates reduces the observed correlations for all three metrics, their relative ordering motivates the weighting adopted in our selection procedure.

\begin{table*}[t]
  \centering
  \small
  \setlength{\tabcolsep}{5pt}
  \begin{tabular}{@{}lrrrrrrrr@{}}
    \toprule
    Selector & Total spans & MQM-wt & Spans/doc & Spans/1k & \%crit & \%clean & Wrong-lang & MetricX$\downarrow$ \\
    \midrule
    Rank fusion & 29,328 & 121,836 & 8.95 & 11.1 & 13 & 14 & 154 & 4.068 \\
    Cross-reference & 29,751 & 125,667 & 9.08 & 11.2 & 13 & 14 & 87 & 4.187 \\
    Best open system & 27,788 & 113,294 & 8.48 & 10.3 & 14 & 16 & 116 & 4.508 \\
    Best closed system & 29,392 & 116,242 & 8.97 & 11.3 & 12 & 17 & 24 & 4.488 \\
    Pooled, 3 metrics & 27,725 & 109,619 & 8.46 & 10.4 & 10 & 16 & 131 & 4.067 \\
    APE-tight, all documents & 30,414 & 129,457 & 9.28 & 11.4 & 15 & 16 & 19 & 4.515 \\
    Per-language APE & 30,261 & 127,653 & 9.23 & 11.4 & 15 & 16 & 20 & 4.519 \\
    C6, language-aware & 27,815 & 108,676 & 8.49 & 10.5 & 11 & 16 & \textbf{0} & \textbf{4.005} \\
    \bottomrule
  \end{tabular}
  \caption{All selections over the 3,277 documents. Rank fusion is Borda
  over MetricX-24 and CometKiwi-XL; cross-reference is cross-system MBR; the pooled
  selector rank-fuses over the six prior selections rather than over the 26; per-language APE uses APE-spans on the CJK pairs and
  APE-tight elsewhere; C6 is the selector of
  Section~\ref{sec:selection}. Span inventory: total; MQM-weighted
  $1{\cdot}$minor$+5{\cdot}$major$+10{\cdot}$critical; per document; per 1k
  characters; percent of documents with a critical span; percent with no spans.}
  
  \label{tab:inventory}
\end{table*}

Across the full set of human-scored WMT25 candidates from 37 systems, all three metrics correlate with human quality at the segment level, with per-pair Pearson correlations ranging from 0.17 to 0.55. However, MetricX is again stronger at the system level, achieving a mean Kendall $\tau$ of 0.29, compared with 0.17 for CometKiwi and 0.12 for xCOMET. The manually defined three-metric fusion reaches 0.26 and therefore does not improve on MetricX alone. ESA scores are not comparable across language pairs, so we fit within each pair and average the resulting coefficients rather than pooling; per-pair ordinary least squares (OLS) gives 0.51 to MetricX, 0.42 to CometKiwi and $-0.07$ to xCOMET, offering no support for weighting xCOMET positively. We test the weightings by leave-one-pair-out cross-validation: weights are fitted on six language pairs and evaluated on the seventh, rotating through all seven. The fitted, hand-set, and equal-weight combinations differ by less than 0.02 Pearson from one another and stay within 0.024 of MetricX alone (Appendix~\ref{app:wmt25}), so the exact weights matter less than not optimising xCOMET. All three metrics reach only 56--58\% segment-level pairwise accuracy against human preference, barely above the 50\% of chance. Individual selections are therefore unreliable, and we base our conclusions on differences that accumulate across all 3,277 documents rather than on any single document being correctly chosen.

\paragraph{Usability of error spans.}
The post-editor consumes xCOMET spans and C6 uses them as a tiebreak, so we checked them against the human MQM-style spans \citep{freitag-etal-2021-experts} on the same WMT25 set. xCOMET flags 6.77 spans per paragraph where a single human annotation pass marks 0.72, a ratio of 9.4, and inflates severities: humans mark these errors minor, xCOMET marks them major or critical. 
Its span counts predict human quality weakly ($r = -0.14$ against ESA, against $-0.69$ for human span counts), and worst on CJK, where over-flagging reaches 39 to 42 times. Across our selectors, total span counts correlate negatively with wrong-language counts ($r = -0.21$): the selectors with fewest spans earned them partly by keeping fluent wrong-language copies that trip no flags. Minimising spans is therefore the wrong objective, and C6 uses them only to arbitrate near-ties and to point the post-editor at suspects.

\section{Results}
\label{sec:results}

Table~\ref{tab:inventory} summarises all selection methods with their span counts and wrong-language counts. Fewer spans does not mean a better reference set: the pooled selector has the fewest spans (27{,}725) but 131 wrong-language documents, while the post-edited output has the most (30{,}414) and none outside the known artefact pair. Editing correct-language text appears to attract flags that fluent wrong-language text does not, so span count alone does not measure pseudo-reference quality. The C-family selectors give the best overall balance, with moderate span counts, no wrong-language selections, and the best MetricX. The best-open setting led by TranslateGemma matches the fused selectors on spans, which says something about current dedicated translation models, but it offers no document-level arbitration and no guarantee of the target language.

\begin{table}[h]
  \centering
  \small
  \resizebox{\columnwidth}{!}{%
  \begin{tabular}{@{}lrrrrr@{}}
    \toprule
    Domain & Docs &  MX & C6-MX &  WL & C6-WL \\
    \midrule
    spoken & 2,090 & \textbf{1.940} & 1.994 & 127 & 0 \\
    social & 737 & 8.483 & \textbf{8.127} & 15 & 0 \\
    news & 259 & 8.986 & \textbf{8.813} & 4 & 0 \\
    software & 98 & 2.589 & \textbf{2.587} & 6 & 0 \\
    edu & 63 & 4.463 & \textbf{4.345} & 2 & 0 \\
    factcheck & 30 & 5.428 & \textbf{5.290} & 0 & 0 \\
    \midrule
    all & 3,277 & 4.068 & \textbf{4.005} & 154 & 0 \\
    \bottomrule
  \end{tabular}}
  \caption{Per-domain comparison of rank fusion and C6. 
  }
  \label{tab:domains}
\end{table}

\subsection{Where the metrics and domains disagree}
\label{sec:domains}

Across domains, C6 improves the mean MetricX score over rank fusion in every case except spoken (Table~\ref{tab:domains}), where 127 of the 154 wrong-language selections occurred: the 0.05 increase in predicted error is the price of replacing fluent Russian with correct Kazakh. Dedicated translation models supply most selected outputs. Our prompt conditions are nested and P4 ran on a different document set, so the P1-to-P4 difference is not a caption effect: on the multimodal subset, where all four ran, GPT-5.5 scores 4.74, 4.45, 4.45 and 4.45 across P1 to P4, and Gemini 4.47, 4.56, 4.59 and 4.57. The analysis stage in P2 accounts for all of GPT-5.5's gain; five-shot and the caption add nothing, and Gemini is helped by none of the three (Appendix~\ref{app:per-pair}).

\subsection{Post-Editing Outcomes}
\label{sec:ape-results}

The stricter post-editor cuts wrong-language output from 154 to 19 overall and to 0 on every genuinely confusable pair, with the best structure fidelity (0.996) and content-anchor preservation (0.911) of the three runs and no change in fluency (Table~\ref{tab:aperuns}, Appendix~\ref{app:ape-behaviour}). MetricX scores APE-tight 0.45 points worse than the rank-fusion selection it edits, but the calibration of Section~\ref{sec:calibration} places both inside the range human-quality translation occupies, and part of the decline reflects the removal of fluent wrong-language output that the neural metrics had preferred. Post-editing edits far more than instructed (Section~\ref{sec:s4}), so we use it as a targeted repair step rather than a pass over every reference: in C6 it provides 29 documents. On the CJK pairs, where xCOMET over-flags the worst, span counts and a native-speaker author's inspection of a sample both favoured APE-spans; that choice is the per-language row of Table~\ref{tab:inventory}, and proper names are kept in their standard target-language forms rather than transliterated.

\subsection{Reference-based \textit{vs.} Human Ranking}
\label{sec:refeval}
Everything above measures the references themselves. For a shared task the question is whether systems scored against them land in the order a human reference would give. We score each participating system with chrF against our references, rank by that score, and compare with the ranking from human ESA~\citep{kocmi-etal-2026-findings}. Over all sixteen pairs, 38 systems agree with the human ranking at Kendall $\tau_b = 0.63$. The informative case is the 1{,}363 segments that also carry a human from-scratch translation, where 30 of them can be ranked against either reference: our references reach $\tau_b = 0.83$ and the human reference $0.82$. The pipeline does produce a reference set as usable for system ranking as the human reference it stands in for. Two caveats. This is chrF only, and systems are annotated on different subsets, so each is scored on its own segments; the comparison between the two references is unaffected, since both use identical data. One asymmetry is worth recording: GPT-5.5 obtains the highest chrF of any system against our references, 64.0, while placing eighth on human ESA, and GPT-5.5 produced 23\% of the reference set. That is the circularity of Section~\ref{sec:discussion}.

\section{Discussion}
\label{sec:discussion}

\paragraph{Which signals to trust.}
The signals available during construction are not equally reliable, and our results order them. Structural checks are exact. Language identification is close, with known confusion pairs to audit by hand. Previous-year human judgments transfer as decisions, not as absolute values. Document-level QE averages are usable as an objective only after the first three have constrained the pool, and error spans belong last, as tiebreaks. The failure we caught sat on the boundary between the second and fourth of these: QE ranked a Russian candidate first, the identifier said it was not Kazakh, and the more reliable signal had to win.

\paragraph{What the human evaluation shows.}
The human evaluation, released after construction, tests that ordering (Appendix~\ref{app:humaneval}, Table~\ref{tab:selcostfull}). Grouped by the family that produced the selected candidate and compared against participating systems on the same segments, our references split in two. Where a frontier-model candidate was kept they are within two ESA points of the strongest systems, and indistinguishable from them on spoken documents; where a translation-specialised open model was kept they fall 14 to 17 points below, and 24 on spoken documents. The anchors barely move between the groups, 82.8 against 82.5 for Gemini~3.1~Pro, so the loss enters at selection rather than in the documents.

\paragraph{Circularity and self-preference.}
Two cautions attach to reading the WMT26 results. On the three pairs where wrong-language selection is most prevalent (\engkaz{}, \engbel{}, \enghye{}) some pools held no trustworthy candidate at all, and the least bad of 26 machine outputs is still a machine output. And the reference-free metrics that built these references are relatives of the reference-based metrics that will consume
them, while our panel plausibly overlaps the GenMT participants and LLM judges prefer their own output over equally good output from others \citep{panickssery-etal-2024-llm}. Three properties limit that exposure without removing it: (1) selection is led by MetricX, a different family from the COMET-family metrics most likely to consume the references; (2) no single system supplies more than 25\% of the references; and (3) every reference carries provenance, so downstream analysis can condition on origin or exclude a family outright. 

\section{Conclusion}
\label{sec:conclusion}
We built pseudo-references for 16 pairs from WMT26 GenMT by
selecting among 80{,}063 machine-translated candidates with reference-free metrics. The setting exposed a failure those metrics do not detect: they score fluent wrong-language output above correct translation, and an unconstrained selector shipped it at scale. A language-identification penalty removes it without cost to the selection objective. Last year's human judgments inform us on which metric to optimise. This year's tell us whether it worked. Systems ranked against our references land close to the order human judges put them in, $\tau_b = 0.83$ against $0.82$ for a human reference. The references match the strongest systems in the task wherever the selector kept a frontier candidate and fall well short wherever it did not. The pipeline is sound; the selection objective is not.


\section*{Limitations}

The largest limitation is what we did not measure. Translation outputs are only spot checked by one human reader for three out of the 16 language pairs, of which only one language pair is briefly checked for translation quality; the other two language pairs are only checked for formatting and output language compliance. Section~\ref{sec:refeval} shows our references reproduce the human system ranking about as well as a human reference does, but on one metric and on the six pairs where a human reference exists. Whether that holds under a neural reference-based metric, and on the ten pairs with no human reference, is untested. Every other quality number in Sections~\ref{sec:gaming} to~\ref{sec:results} comes from the same reference-free metrics that built the references, which is circular by construction. The objective and the repair prompt were calibrated on seven of the 16 pairs; for the other nine, including every pair where we found wrong language selection, that calibration is an extrapolation. CometKiwi and xCOMET share the COMET lineage, so their agreement is weaker evidence than it appears. The structure gate checks element counts, not content placement within them. Closed-model candidates are not reproducible.


\section*{Environmental Impact Statement}
We present an audit of the carbon emission for parts of the experiments in this work. Although the audit only covered the experiments involving GPT-5.5 and not the complete experiments, we think that the partial information is still valuable as a reference for estimating the environmental impact of different machine translation settings using LLMs.  

The accumulated token usage and carbon footprint of GPT-5.5 is reported by the service provider\footnote{\url{https://learn.microsoft.com/en-us/power-bi/connect-data/azure-emissions-calculation-methodology}}. Table \ref{tab:carbonfootprint} showed the total number of input, reasoning, output tokens and an estimation of the carbon footprint of each GPT-5.5 cell.

\begin{table}
    \centering
    \begin{tabular}{l|rr}
      Cell   & \#tokens & emission (kgCO2e)\\
      \hline
      p1   & 3.13M & 0.06\\
      p2   & 5.17M & 0.09\\
      p3   & 9.78M & 0.17\\
      p4-gemini   & 8.43M & 0.15\\
      p4-qwen   & 8.57M & 0.15\\
      APE-scores & 7.44M & 0.13\\
      APE-spans & 8.94M & 0.16\\
      APE-tight & 14.57M & 0.26\\
      \hline
      total   & 66.03M & 1.18\\
    \end{tabular}
    \caption{Total number of input, reasoning, output tokens and an estimation of the carbon footprint of each GPT-5.5 cell}
    \label{tab:carbonfootprint}
\end{table}


\section*{Acknowledgements}
The computation cost for the experiments involving GPT-5.5 is provided by the National Research Council of Canada. 

\bibliography{custom,anthology-1,anthology-2}

\appendix

\section{Related Work}
\label{sec:related}
\paragraph{Pseudo-references.}
Using machine translations in place of human references goes back to \citet{albrecht-hwa-2007-regression, albrecht-hwa-2008-role}, who trained sentence-level evaluation regressors on pseudo-references from off-the-shelf systems. 
Reference quality is believed to bound evaluation quality, but inconclusive evidence is presented over the years. \citet{freitag-etal-2020-bleu} showed that standard human references reward translationese, and that rankings by automatic evaluation metrics shift when paraphrased references are used instead. Subsequently, \citet{freitag-etal-2023-results} found that low-quality human references can dramatically hurt the performance and reliability of evaluation metrics. WMT findings have repeatedly reported reference translations losing to top systems in human evaluation \citep{kocmi-etal-2025-findings}. On the other hand, \citet{mathur-etal-2020-results} found that the variations of metrics' correlation with humans are minimal when they score translations against references with different quality. 

Pseudo-references, which by nature are MT output, inherit the potential quality problems from whichever systems and metrics build them, which is why we spend half of this paper on what the selection metrics get wrong and our attempts to add quality control measures for generating pseudo-references of better quality.

\paragraph{Metric-guided selection and its failure modes.}
Choosing among candidate translations by a neural utility function is the basis of minimum Bayes risk (MBR) decoding \citep{eikema-aziz-2020-map, freitag-etal-2022-high}, which scores each candidate against the others and keeps the one with the highest expected utility, and of reranking under quality-aware decoding \citep{fernandes-etal-2022-quality}. Our cross-reference selector is cross-system MBR with a COMET utility \citep{rei-etal-2020-comet}, and we claim no novelty in the selection arithmetic. The relevant precedent for our findings is \citet{amrhein-sennrich-2022-identifying}, who used MBR as a probe to surface COMET's blind spots. Given a large enough candidate pool, the metric selects untranslated copies of the source and translations with corrupted numbers. We encounter the same failure without setting out to find it. 
With 26 heterogeneous candidate systems, reference-free metrics promote fluent wrong-language output, and the selector needs an explicit language constraint the metrics do not provide. 
\citet{kocmi-etal-2021-ship} showed that metric decisions disagree with humans on close system pairs, and our WMT25 calibration extends this to the selection regime, where all candidates are near the top of the quality range. 

Similarly, \citet{yan-etal-2023-bleurt} discovered another failure mode by guiding MBR using BLEURT \citep{sellam-etal-2020-bleurt}. BLEURT would give high scores to some translation outputs, which they called universal translations, that have nothing to do with the source or the reference translation. Since we did not use BLEURT as our selector and did not observe such universal translations phenomenon in the spot checks of the MT output, we did not add further measures to mitigate this metric-guided selection failure mode.  

\paragraph{Language confusion.}
Translating into the wrong language is a documented failure of  multilingual systems, both as off-target translation in massively multilingual NMT \citep{zhang-etal-2020-improving} and as language confusion in instruction-tuned LLMs \citep{marchisio-etal-2024-understanding}. Our claim is not that off-target output exists but that reference-free selection actively prefers it. \citet{liu-etal-2026-mending} show the same vulnerability on the training side,
where multilingual RL reward models can be hacked by off-target output. Our claim concerns inference: reference-free selection does not merely tolerate off-target output, it prefers it. Presented with a pool that contains a fluent off-target candidate, the metrics that are supposed to identify the best translation promote it above correct ones. Any pipeline that optimises a reference-free score without an explicit language constraint will therefore concentrate off-target text in the lower-resource pairs, where the reference is the only available anchor and a corrupted one is most damaging.

\paragraph{Reference-free QE and LLM evaluators.}
We score with MetricX-24 \citep{juraska-etal-2024-metricx}, CometKiwi \citep{rei-etal-2022-cometkiwi, rei-etal-2023-scaling}, and xCOMET \citep{guerreiro-etal-2024-xcomet}, three systems from the top of recent WMT metrics and QE tasks \citep{freitag-etal-2024-llms}. LLMs also judge translation quality directly \citep{kocmi-federmann-2023-large, zheng-etal-2023-judging}, and judge-model self-preference is documented \citep{panickssery-etal-2024-llm}. Self-preference matters here because our candidate systems plausibly overlap with GenMT participants: a reference selected by metrics that favour a model family inflates that family's evaluation. We discuss this circularity in Section~\ref{sec:discussion}. For repair we use LLM-based automatic post-editing, following evidence that GPT-4-class models post-edit conservatively when instructed \citep{raunak-etal-2023-leveraging}.

\section{Panel Validation}
\label{app:panel}

Table~\ref{tab:usable} reports the per-cell validation of the candidate panel before scoring. A document is usable if a real target-language translation is present. Misses split into wrong-script output, malformed output (placeholders, truncation), source copies, and, for
the analysis-style prompts, leaked reasoning blocks. For each cell we report the dominant failure mode with its count and aggregate the remaining misses under \emph{other}.  

\begin{table}[h]
  \centering
  \resizebox{\columnwidth}{!}{%
  \begin{tabular}{@{}llrrlr@{}}
    \toprule
    Model & Cell & Total & Usable & Dominant failure & Other \\
    \midrule
        GPT-5.5 & p1 & 3,277 & 3,277 & none & 0 \\
     & p2 & 3,277 & 3,277 & none & 0 \\
     & p3 & 3,277 & 3,277 & none & 0 \\
     & p4-qwen & 2,706 & 2,706 & none & 0 \\
     & p4-gemini & 2,706 & 2,702 & malformed - 4 & 0 \\
    Gemini-3.1 & p1 & 3,277 & 3,276 & malformed - 1 & 0 \\
     & p2 & 3,277 & 3,277 & none & 0 \\
     & p3 & 3,277 & 3,277 & none & 0 \\
     & p4-gemini & 2,706 & 2,706 & none & 0 \\
     & p4-qwen & 2,706 & 2,705 & malformed - 1 & 0 \\
    TranslateGemma & native & 3,277 & 3,277 & none & 0 \\
    HY-MT1.5 & native & 3,277 & 2,873 & wrong script - 375 & 29 \\
    Aya-32B & p1 & 3,277 & 3,277 & none & 0 \\
     & p2 & 3,277 & 3,141 & analysis leak - 119 & 17 \\
     & p3 & 3,277 & 3,150 & analysis leak - 109 & 18 \\
     & p4-qwen & 2,706 & 2,569 & analysis leak - 118 & 19 \\
     & p4-gemini & 2,706 & 2,554 & analysis leak - 136 & 16 \\
    Tower-Plus & p1 & 3,277 & 3,242 & wrong script - 34 & 1 \\
     & p2 & 3,277 & 3,232 & wrong script - 26 & 19 \\
     & p3 & 3,277 & 3,234 & wrong script - 27 & 16 \\
     & p4-qwen & 2,706 & 2,668 & wrong script - 29 & 9 \\
     & p4-gemini & 2,706 & 2,676 & wrong script - 24 & 6 \\
    EuroLLM-22B & p1 & 3,277 & 3,209 & wrong script - 50 & 18 \\
     & p2 & 3,277 & 2,754 & wrong script - 453 & 70 \\
     & p3 & 3,277 & 2,761 & wrong script - 446 & 70 \\
     & p4-qwen & 2,706 & 2,219 & wrong script - 442 & 45 \\
    \bottomrule
  \end{tabular}}
 \caption{Panel usability per cell. HY-MT's wrong-script output concentrates on en--hy (188) and en--kk (187), scripts it cannot produce. Aya's analysis prompts copy the English reasoning block on hard pairs. EuroLLM fails non-Latin targets.}
  \label{tab:usable}
\end{table}

HY-MT cannot produce Armenian or Kazakh script and emits mixed-script output on those pairs (404 unusable documents, 375 of them wrong-script). EuroLLM loses 487 to 523 documents per analysis-prompt cell on non-Latin targets, predominantly to wrong-script output (442 to 453 per cell). Aya emits its English analysis block instead of the translation on hard pairs under P2--P4 (109 to 136 documents per cell, of 127 to 152 unusable in total) while its P1 cell is complete.

\section{Editing Behaviour of the APE Prompts}
\label{app:ape-behaviour}

\begin{table}[h]
  \centering
  \resizebox{\columnwidth}{!}{%
  \begin{tabular}{@{}lcccc@{}}
    \toprule
     & Input (\#1) & APE-scores & APE-spans & APE-tight \\
    \midrule
    Language (langid, wrong docs) & 154 & 21 & 24 & 19 \\
    Structure fidelity & 0.987 & 0.995 & 0.995 & \textbf{0.996} \\
    Content anchors & 0.863 & 0.839 & 0.834 & \textbf{0.911} \\
    Fluency & 0.971 & 0.971 & 0.971 & 0.971 \\
    Median similarity to input & -- & 0.590 & 0.562 & 0.571 \\
    Documents byte-identical to input & -- & 15 & 3 & \textbf{47} \\
    MetricX mean $\downarrow$ & \textbf{4.068} & 4.535 & 4.585 & 4.515 \\
    \bottomrule
  \end{tabular}}
  \caption{The three post-editing runs over all 3,277 documents. APE-scores sees QE scores, APE-spans adds xCOMET spans, APE-tight is the tightened prompt. Structure fidelity, content-anchor preservation, and fluency are the heuristic measures of Appendix~\ref{app:stats}. The language row for the runs uses the same langid audit as Table~\ref{tab:wronglang} computed on the APE outputs directly (run counts include the en--id artefact).}
  \label{tab:aperuns}
\end{table}

APE-tight wins on every signal we trust (language, structure, anchors, restraint) and loses only on neural QE, which Section~\ref{sec:calibration} discounts at this score range.
The per-document rate at which APE-tight beats the unedited input is 21.9\% on MetricX, 24.0\% on CometKiwi, and 31.6\% on xCOMET, with the rest ties or losses: on text that is already at the top of the range, post-editing is as likely to move a score down as up, which is the expected behaviour of noisy metrics near their ceiling rather than evidence of harm.

\begin{table*}[t]
  \centering
  \resizebox{\textwidth}{!}{%
  \begin{tabular}{@{}lrrrrrrrrr@{}}
    \toprule
    Pair & TransGemma & HY-MT & GPT5 p1 & GPT5 p4q & Gemini p1 & Gemini p4q & Aya p1 & Tower p1 & EuroLLM p1 \\
    \midrule
    \cesdeu{} & 4.07 & 4.42 & 4.44 & \textbf{2.69} & 4.52 & 2.73 & 4.47 & 4.61 & 8.69 \\
    \engbel{} & \textbf{5.32} & 5.51 & 6.29 & 5.60 & 6.05 & 5.81 & 12.12 & 8.50 & 12.59 \\
    \engdeu{} & \textbf{2.54} & 3.00 & 3.31 & 2.62 & 3.21 & 2.89 & 3.10 & 3.15 & 3.35 \\
    \engekk{} & \textbf{6.14} & 6.47 & 6.97 & 6.41 & 7.06 & 6.76 & 15.31 & 14.49 & 7.53 \\
    \enghye{} & 5.28 & 6.36 & 5.33 & \textbf{4.41} & 4.90 & 4.65 & 19.01 & 17.78 & 18.76 \\
    \engind{} & \textbf{3.27} & 3.57 & 4.41 & 3.68 & 4.08 & 3.77 & 4.36 & 6.13 & 8.50 \\
    \engisl{} & 6.13 & 6.10 & 6.16 & \textbf{5.69} & 6.35 & 6.11 & 14.85 & 6.13 & 16.90 \\
    \engjpn{} & \textbf{4.38} & 4.52 & 5.09 & 4.39 & 4.67 & 4.52 & 5.09 & 4.85 & 5.21 \\
    \engkaz{} & 5.55 & 11.21 & 6.10 & \textbf{5.37} & 5.63 & 5.41 & 15.74 & 11.00 & 15.62 \\
    \engkor{} & \textbf{4.28} & 4.42 & 5.04 & 4.57 & 4.75 & 4.45 & 4.96 & 5.05 & 5.34 \\
    \engrus{} & \textbf{3.72} & 4.11 & 5.15 & 4.18 & 4.99 & 4.62 & 5.03 & 4.83 & 5.12 \\
    \engtha{} & \textbf{3.75} & 4.01 & 4.55 & 3.95 & 4.33 & 3.90 & 9.26 & 7.39 & 15.84 \\
    \engukr{} & \textbf{4.46} & 4.67 & 5.60 & 5.03 & 5.60 & 5.32 & 5.69 & 5.41 & 5.95 \\
    \engzhs{} & 3.19 & 3.19 & 3.84 & \textbf{3.11} & 3.55 & 3.16 & 3.91 & 3.61 & 3.65 \\
    \engzht{} & \textbf{3.25} & 3.44 & 3.91 & 3.32 & 3.67 & 3.28 & 3.88 & 3.70 & 3.88 \\
    \zhsjpn{} & 5.54 & 6.28 & 6.77 & 5.64 & 6.24 & \textbf{5.42} & 6.27 & 6.22 & 6.61 \\
    \midrule
    all pairs & \textbf{4.42} & 5.05 & 5.16 & 4.45 & 4.96 & 4.58 & 8.18 & 6.97 & 8.97 \\
    \bottomrule
  \end{tabular}}
  \caption{Candidate-level MetricX means (lower is better) per pair for nine of the 26 cells: the two native cells, the P1 and P4-qwen cells of the two closed models, and the P1 cells of Aya, Tower-Plus and EuroLLM, their best condition. Bold: best of the shown cells, except that the P4-qwen columns are computed on that model's 2,706-document multimodal subset and are not comparable with the rest.}
  \label{tab:s2pairs}
\end{table*}

\section{Per-Pair Results}
\label{app:per-pair}

Table~\ref{tab:s2pairs} gives the candidate-level MetricX means per pair for a representative subset of cells, and the best cell per pair under each metric. Two regularities hold across the full 26-cell matrix (available with the released artefacts). The dedicated translation models are the panel backbone: TranslateGemma has the best or near-best mean on most pairs. Restricted to the multimodal subset, where the comparison is valid, the caption cells are indistinguishable from their five-shot counterparts: GPT-5.5 P4 differs from P3 by 0.002 MetricX and Gemini P4 from P3 by 0.022. TranslateGemma remains the best cell on 11 of the 16 pairs.

Rank-fusion selection counts mirror the means: TranslateGemma and HY-MT are the two most-selected systems on 13 of 16 pairs under the unconstrained selector, with GPT-5.5 and Gemini prompt cells taking most of the remainder, concentrated in news, education, and software. Under C6 the balance shifts toward the closed models on the penalised pairs, since their candidates are language-correct more often on Kazakh, Belarusian, and Armenian.

\section{Worked Example: \engkaz{} Document 24409}
\label{app:example}

The source is a spoken-domain transcript (Figure~\ref{fig:ex-src}). The
organisers' per-row instruction names the target language explicitly, opening
``You are a professional Kazakh translator'' and closing ``Please translate the
following text into Kazakh (kaz\_Cyrl)'', so the wrong-language candidate does
not arise from an ambiguous request.

\begin{figure}[h]
{\footnotesize
\begin{verbatim}
Hello?
Yes, Lois.
I'll hold the wire, please.
Lois, phone.
Take the message, will you, please?
Dad.
She's busy right now.
I'll take the message.
Uh, what was that?
Ralph is calling for instead of Henry.
What?
But he can't meet her in front of the gymnasium.
Can she?
Go where?
Uh, well, wait a minute.
\end{verbatim}
}
\caption{The source document for \texttt{24409-kaz\_Cyrl}: an ASR transcript of
mid-century film dialogue, one utterance per line, which the translation must
preserve.}
\label{fig:ex-src}
\end{figure}

Table~\ref{tab:example} gives the two candidates that decide the selection: the
pool's best-scoring candidate and the one C6 chooses.

\begin{table}[t]
  \centering \footnotesize \setlength{\tabcolsep}{2.5pt}
  \begin{tabular}{@{}lrrrlrr@{}}
    \toprule
    Candidate & MX$\downarrow$ & CK & xC & langid ($p$) & $F$ & $F-\lambda p$ \\
    \midrule
    eurollm-p3 & \textbf{1.65} & 0.772 & 0.945 & ru (1.00) & \textbf{0.82} & $-2.18$ \\
    gemini-p4q & 2.58 & 0.720 & 0.890 & kk (1.00) & 0.71 & \textbf{0.71} \\
    \bottomrule
  \end{tabular}
  \caption{The EuroLLM candidate is fluent Russian and best on all three
  metrics, so no reweighting would have rejected it; the language penalty
  demotes it below every correct-Kazakh candidate.}
  \label{tab:example}
\end{table}

Under C6 the Russian candidate's MetricX of 1.65 standardises to $F = 0.82$, the
top fused score in the pool ($\mu = -8.42$, $\sigma = 8.21$ on oriented
MetricX). The identifier labels it Russian at confidence 1.0, so the penalty
subtracts $\lambda p = 3.0$ and takes it to $-2.18$, below every
correct-language candidate. The Kazakh candidate that scored 2.58 wins at
$F = 0.71$.

The document is also multimodal: the source is a video
(\texttt{media\_00130.mp4}), and the P4 cells receive a caption of it rather
than the media itself. Both captioners identify the speakers and the named
entities \emph{Lois}, \emph{Ralph} and \emph{Henry}, and the winning candidate
comes from a P4 cell.

\section{WMT25 Meta-Evaluation Detail} 
\label{app:wmt25}

\paragraph{How well do xCOMET spans locate human-annotated errors?}
The comparison uses the 542 human-best documents carrying at least one human
span. Matching by character offset, xCOMET recovers 45.8\% of human spans
(859 of 1,874) and 19.8\% of its own flags land on one (878 of 4,433). The two
counts differ because several xCOMET flags can fall inside a single human span.

\paragraph{How much does xCOMET over-flag, by language?}
On the 542 documents with at least one human span, xCOMET produces 41.7 spans
per human span on en--zh and 39.2 on en--ja, an order of magnitude above the
Latin-script pairs (cs--de 14.0, en--ru 10.0, en--is 4.2, en--et 3.7), with
en--uk intermediate at 31.2. These ratios are computed on a different base from the 9.4$\times$ figure in Section~\ref{sec:calibration}, which averages over all
1,409 paragraphs.

\paragraph{Do span counts track human error counts?}
Not usefully. xCOMET's minor-span counts correlate \emph{negatively} with human
span counts in all seven pairs: where humans mark more errors, xCOMET marks
fewer minor ones. Total-span counts are inconsistent rather than merely weak,
ranging from $-0.08$ (en--zh) to 0.43 (en--ru).

\paragraph{Do the fusion weights matter?}
Leave-one-pair-out fitting gives held-out Pearson correlations of 0.370 for per-fold OLS, 0.362 for the hand set weights, 0.358 for equal weights, and 0.346 for MetricX alone. The spread of 0.024 across all four is small, which is why Section~\ref{sec:calibration} concludes that the exact weights matter less
than which metric leads.

\section{Statistical Methods and Heuristic Measures}
\label{app:stats}
Significance for paired selector comparisons uses document-level differences ($n = 3{,}277$): percentile bootstrap over documents (2,000 resamples) for confidence intervals, two-sided sign-flip permutation (2,000 permutations) for $p$-values, and Cohen's $d$ as mean over standard deviation of the differences. The penalty saturates at $\lambda = 3$: wrong-language documents fall from 162 at $\lambda = 0$ to 18 at $0.5$, 12 at $1$, 5 at $2$, and 0 at $\lambda \geq 3$. Correlations are Pearson, with Spearman as Pearson on ranks. MetricX is sign-flipped so positive means agreement. The MQM span weighting is $1 \cdot \text{minor} + 5 \cdot \text{major} + 10 \cdot \text{critical}$.

The heuristic measures used for post-edit evaluation are: \textit{structure fidelity (SF)}, agreement of paragraph, line, and markup-element counts between source and output; \textit{content-anchor preservation (CP)}, the fraction of source URLs, handles, hashtags, and numbers present in the output; and \textit{fluency (FL)}, a distinct-4-gram ratio computed over words for spaced scripts and characters otherwise. A fourth measure, \textit{script-ratio language correctness (LC)}, over-counted wrong language on kanji-only Japanese and on untranslated proper names; its findings were retracted and replaced by the langid audit of Section~\ref{sec:gaming}, and it is not used in this paper's claims.

\section{Reproducibility}
\label{app:repro}

Every run in the pipeline is a configuration file. Every stage reads and writes JSONL, and every selected reference carries a provenance record: producing system and prompt cell, the candidate pool with scores, gate outcomes, selector arithmetic, language identification labels, and the post-editing outcome where applied. The selection variants, span inventories, meta-evaluation scripts, and the langid audit are single scripts over these artefacts. Every number in the results tables was recomputed from the raw artifacts during writing: wrong-language counts by re-running the language audit on the submission files; selector score means, span inventories, composition, and significance tests by matching every selected document back to its producing cell and its stored scores; the WMT25 correlations, rankings, fitted weights, and cross-validation by independent reimplementation over the score and judgment archives. The verification scripts and their output accompany the paper sources. 


\section{Generation Prompt Templates (P1--P4)}
\label{app:gen-prompts}

All four generation conditions share one instruction spine and differ
only in which optional blocks are switched on. Table~\ref{tab:panel}
gives which models run which conditions.

\paragraph{P1 (template \texttt{T0}).}
P1 adds no template of our own: the organiser's per-row instruction is
passed as the system message and the source document as the user
message, so that P1-vs-P2 differences are attributable to our additions
and never to plumbing.

{\scriptsize
\begin{verbatim}
[system] <<INSTRUCTION>>
  # the WMT26 per-row organiser instruction,
  # released with the blindset, byte-for-byte
[user]   <<SOURCE_DOC>>
  # the source document, whole
\end{verbatim}
}

\paragraph{P2--P4 (analysis spine, \texttt{T2}).}
P2, P3 and P4 all render the template below, wrapping the same
\texttt{<<INSTRUCTION>>} used in P1. Slots are filled by token
replacement, never \texttt{str.format}, so JSON and brace content in the
source is safe, and empty slots are dropped before the call.
\texttt{<<SCRIPT\_NOTE>>} and \texttt{<<LP\_NOTE>>} are per-pair
one-line guides (e.g.\ ``Write the translation in Cyrillic script.'', or
a regional-variety note).
\texttt{<<FEW\_SHOT>>} is empty for P2 and holds the five-shot block for
P3 and P4. The caption block is appended for P4 only.

{\scriptsize
\begin{verbatim}
You are an expert <<TGT_NAME>> translator and
translation analyst.

Follow this instruction exactly. It is graded and
will be shown to the human evaluators; not following
it counts as a translation error. If it conflicts
with the guidance below, THE INSTRUCTION WINS:

<<INSTRUCTION>>

Work in two stages.

STAGE 1 - Interpret the source (write this analysis
in English, inside <analysis> tags, concise bullet
points):
1. INSTRUCTION: restate the specific constraints
   this instruction imposes (formality/voice,
   glossary or terminology, output format,
   morphology, style) and how each will be
   satisfied.
2. MEANING RISKS: idioms, irony/sarcasm, ambiguous
   words, culture-specific items, ellipsis needing
   context - state the intended meaning of each.
3. FORM RISKS in <<TGT_NAME>>: formality (T-V),
   grammatical gender agreement, honorifics,
   plurals, word order around markup/placeholders -
   decide each.
4. INVARIANTS: what must survive exactly - names,
   numbers, tags, URLs, @handles, placeholders,
   line/paragraph structure.

STAGE 2 - Translate, applying every decision from
Stage 1. Preserve all source structure and markup;
translate only human-readable text; the result must
read as natural <<TGT_NAME>>, and the analysis must
NOT appear in it.
<<SCRIPT_NOTE>>
<<LP_NOTE>>
<<DOMAIN_NOTE>>
<<FEW_SHOT>>
Output format, exactly:
<analysis>...</analysis>
<t>...the translated document...</t>
\end{verbatim}
}

\paragraph{Domain modules (\texttt{<<DOMAIN\_NOTE>>}).}
One module is selected per document by domain and substituted into the
spine. The two conversational domains are given first.

{\scriptsize
\begin{verbatim}
--- spoken -----------------------------------------
This is an automatic speech-recognition transcript,
one utterance per line, and may contain recognition
errors. Translate it as natural spoken <<TGT_NAME>>:
- Keep the same number of lines and the same line
  order; do not merge or split lines.
- Render it as speech, not polished prose: keep
  interjections; drop non-linguistic sounds
  (laughter, hesitation noises); if a word is cut
  off, infer the full word or omit it; keep foreign
  words as they are.
- The original video is the human-evaluation
  reference, so favour what the speaker most
  plausibly meant.

--- social -----------------------------------------
This is a social-media post given as HTML. Translate
only the human-visible text:
- Keep every HTML tag and attribute, @mentions, URLs
  and emoji exactly as they appear; never translate
  or reorder them.
- Preserve the blank-line separation between blocks
  exactly (the same number of blank-line-separated
  blocks in and out).
- Match an informal, conversational register; render
  expressive style (capitalisation, character
  elongation, expressive punctuation) naturally in
  <<TGT_NAME>> rather than flattening it.
- Decode HTML entities the way the source uses them;
  do not introduce new tags.
\end{verbatim}
}

The remaining modules cover the text-heavy and structured domains, with
a fallback for anything unmatched.

{\scriptsize
\begin{verbatim}
--- news / factchecking ----------------------------
This is a news or fact-checking article (plain text,
or HTML):
- Keep the paragraph and blank-line structure
  exactly (the same number of blank-line-separated
  paragraphs), and preserve any HTML tags and
  non-breaking spaces unchanged.
- Translate the headline (first line) using
  <<TGT_NAME>> headline conventions.
- Names, titles, figures, dates, amounts and
  citations stay exact; translate direct quotations
  faithfully using <<TGT_NAME>> quotation
  punctuation.
- Use the register of quality <<TGT_NAME>>
  journalism; report claims neutrally.

--- software ---------------------------------------
This is software / UI localisation content, and may
be a JSON object or array (sometimes inside a code
fence):
- If the source is JSON, return valid JSON of the
  same shape and length: translate the values only,
  never the keys.
- Copy templating placeholders (double-curly tokens
  such as the count placeholder) and any markup
  byte-identical, arranging <<TGT_NAME>> word order
  naturally around them.
- Keep UI strings short enough for buttons and
  labels; keep marketing strings punchy.

--- general (fallback) -----------------------------
Translate faithfully and idiomatically into
<<TGT_NAME>>. Preserve any markup, and keep all
numbers, names, URLs and placeholders exact.
\end{verbatim}
}

\paragraph{Five-shot block (\texttt{<<FEW\_SHOT>>}, P3 and P4).}
The slot is filled with up to five post-edited examples for the
canonicalised language pair.

{\scriptsize
\begin{verbatim}
Examples:
Example 1
Source: <<SRC_EXAMPLE_1>>
Translation: <<TGT_EXAMPLE_1>>
Example 2
Source: <<SRC_EXAMPLE_2>>
Translation: <<TGT_EXAMPLE_2>>
... (up to 5)
\end{verbatim}
}

\paragraph{Caption block (P4 only).}
P4 runs on the 2{,}706-document multimodal subset, with captions
produced by either Qwen3-Omni or Gemini-3.5-Flash (the two P4 cells).
The model receives the caption text only, never the media itself.

{\scriptsize
\begin{verbatim}
Visual context (from the <source video | source
social-media image(s)>, to complement the text;
translate the content, not this note):
<<CAPTION>>
\end{verbatim}
}

\paragraph{Native templates.}
\label{app:native-prompts}
The two dedicated translation models run a single pass under their own
native templates, with no analysis stage, five-shot block, or caption.

{\scriptsize
\begin{verbatim}
--- TranslateGemma-27B ---
Translate from {src_bcp47} to {tgt_bcp47}:
{src_text}

--- HY-MT1.5-7B ---
Translate the following segment into {tgt_name},
without additional explanation.
{src_text}
\end{verbatim}
}


\section{The APE-scores and APE-spans Prompts}
\label{app:ape-scores-spans}

These two post-editing prompts share the template below and differ
only in whether xCOMET error spans are shown: APE-scores gives the
MetricX, CometKiwi and xCOMET document scores, and APE-spans
additionally lists each candidate's error spans. Lines marked
\texttt{[spans only]} appear in APE-spans but not in
APE-scores. Both were run at reasoning\_effort=medium and
post-edit the rank-fusion selection over the top three candidates, which
always include the cross-reference winner.

{\scriptsize
\begin{verbatim}
You are an expert {TGT} translator and post-editor.
Produce the single best reference translation (a
"pseudo-reference") of the SOURCE for a
machine-translation evaluation: maximally faithful
to the source, fluent and idiomatic, and internally
coherent as a whole document. 
Source language:{SRC}. 
Target language: {TGT}. 
Domain: {DOMAIN}.
SOURCE: {SOURCE_DOC}
You are given the best candidate translations,
pre-ranked by two independent quality-estimation
systems:
- MetricX-24: error score 0-25, LOWER is better.
- CometKiwi-XL: quality 0-1, HIGHER is better.
- xCOMET-XL: quality 0-1, HIGHER is better.
  [spans only: "...HIGHER is better (with flagged
   error spans below each candidate)."] Selector
   opinions: rank-fusion (combining both metrics)
   prefers [{RF_WINNER}]; cross-reference consensus
   prefers [{XR_WINNER}] (they AGREE. | they
   DISAGREE - judge carefully.)
CANDIDATES (ranked best->worst):
[{rank}] {system} | MetricX {mx} |
         CometKiwi {ck} | xCOMET {xc}
{candidate_translation}
  [spans only] xCOMET flagged errors (minor={a},
   major={b}, critical={c}):
  [severity] 'span text' ; [severity] 'span text' ;
... (repeated for the top-3 candidates)

INSTRUCTIONS. Reason carefully (internally): judge
each candidate against the SOURCE for ADEQUACY
(every meaning preserved; nothing added, dropped, or
hallucinated) and for FLUENCY, weighing the QE
scores and BOTH selector opinions[spans only: ", and
the flagged xCOMET error spans"]. Decide whether to
keep the strongest candidate as-is, post-edit it, or
merge the best parts of several; if you merge, keep
the result COHERENT - one consistent terminology,
register and style throughout, with no
contradictions or visible seams. Then output ONLY
the single best {TGT} translation of the WHOLE
source: faithful, natural, and coherent end-to-end,
preserving the source's structure and any markup
EXACTLY (same segment/line/paragraph count; keep
HTML/JSON tags intact). Wrap it in <t></t> and put
NO reasoning or commentary in the output.
\end{verbatim}
}

\section{The Tightened Post-Editing Prompt}
\label{app:ape-prompt}

The APE-tight prompt has five blocks: role and task; a hard
language rule with a pair-specific trap line; the source document; the
evidence packet with instructions on how to read it; and a procedure
with an output contract. Slots in braces are filled at run time. The
\texttt{\{PAIR\_TRAP\}} slot names the confusable neighbour to avoid for
each target: Kazakh not Russian, Belarusian not Russian, Ukrainian not
Russian, Armenian script not Cyrillic, Japanese with kana not a
Chinese-character-only copy, and Korean Hangul. The prompt was run at
\texttt{reasoning\_effort=high} with a 128k output-token ceiling.

{\scriptsize
\begin{verbatim}
You are an expert {TGT} translator and post-editor.
Produce ONE final reference translation (a
"pseudo-reference") of the SOURCE document for
machine-translation evaluation. It must be written
entirely in {TGT} using {SCRIPT}, be maximally
faithful to the source, fluent and idiomatic, and
coherent as a whole document.

Source language: {SRC}. 
Target language: {TGT}
({SCRIPT}). 
Domain: {DOMAIN}.

HARD LANGUAGE RULE. Every segment of your output
must be in {TGT} ({SCRIPT}). Do not leave, copy, or
emit text in any other language, except for numbers,
dates, units, currency, URLs, emails, @handles, and
code, even if a candidate below is in another
language and reads fluently. {PAIR_TRAP}. A
candidate can score well on the quality metrics
while being in the WRONG language: the metrics are
fooled by fluent non-translations. If any candidate
is wholly or partly not in {TGT}, discard that
candidate (or that part) whatever its scores, and
translate from the SOURCE.

SOURCE:
{SOURCE_DOC}

You are given the strongest candidate translations,
each with quality-estimation signals and, where
flagged, xCOMET error spans:
- MetricX-24: error 0-25, LOWER better.
- CometKiwi-XL: 0-1, HIGHER better.
- xCOMET-XL: 0-1, HIGHER better, with flagged spans.

HOW TO READ THESE. Each score is computed per unit
(segment/paragraph) and does NOT see the whole
document. The scores and spans are noisy and can be
wrong in either direction. Use them ONLY as pointers
to where a problem may be. Do not trust any single
score. For each candidate, combine all three scores
and the spans to judge one thing: how erroneous is
this candidate as a WHOLE document, after you have
checked the flagged places against the SOURCE
yourself.

Selector signals (these are only signals, not ground
truth): rank-fusion prefers [X]; cross-reference
prefers [Y]; (they agree|they disagree). Weigh them,
but do not let them override the language rule or
your own check against the SOURCE.

CANDIDATES (ranked best->worst):
{CANDIDATES_WITH_SCORES_AND_SPANS}

PROCEDURE (reason internally; your reasoning must be
cogent; output none of this):
1. Apply the language rule: discard any candidate
   not fully in {TGT}/{SCRIPT}, whatever its scores.
2. Among the rest, pick the single least-erroneous
   candidate at document level as your BASE.
3. Verify each flagged span and each score-implied
   problem against the SOURCE. Keep a segment
   EXACTLY as in the base when it is already
   correct; change only the specific spans you have
   confirmed wrong. Do not paraphrase, re-order,
   restyle, or swap synonyms in segments that are
   already correct. Prefer minimal edits to one base
   over merging; borrow a phrase from another
   candidate only when it is clearly and verifiably
   better, keeping terminology, register and style
   consistent with no visible seams.
4. Preserve exactly, never translate or alter:
   numbers, dates, units, currency, URLs, emails,
   @handles, and code that belong in source form.
5. Preserve the source structure and markup EXACTLY:
   same segment/line/paragraph count; keep every
   HTML/JSON tag intact.

OUTPUT. Return ONLY the one final {TGT} translation
of the WHOLE source, entirely in {TGT} ({SCRIPT}),
faithful, natural, coherent end-to-end, structure
and markup preserved exactly. Wrap it in <t></t>.
Put no reasoning, notes, labels, candidate names,
alternatives, or commentary anywhere in the output.
\end{verbatim}
}

\section{Post-hoc Validation on the WMT26 Human Evaluation}
\label{app:humaneval}
The WMT26 GenMT human evaluation~\citep{kocmi-etal-2026-findings} scores our reference set as a participating system under the name \texttt{PseudoRef}. All figures below are recomputed from that release. Annotation items are joined to our documents through the organisers' identifier crosswalk on $(\textrm{domain}, \textrm{true\_doc\_id}, \textrm{src\_lang}, \textrm{tgt\_lang})$, which is one-to-one over our 3,277 documents and resolves every in-scope item. The identification of our system rests on exact reconstruction: the annotated segments, concatenated in index order, reproduce our C6 submission character for character on 2,292 of 2,292 documents, where the next closest of the fourteen selection files we built reaches 94.8\%. Every number in this appendix was produced twice, by two independently written pipelines over the raw release.

\paragraph{What the release contains.}
Twenty-three campaigns, 102,060 annotation items outside the tutorial and attention-check pools, and 561,481 error spans over 48 systems, of which 44 appear in our sixteen pairs. Scoring is Error Span Annotation \citep{kocmi-etal-2024-error}: annotators mark character spans and grade each
minor or major, and the release carries no error typology, the category field being empty on all 561,481 spans. Severities split 49.1\% major to 50.9\% minor. Where a segment and system pair carries more than one annotation pass, which happens for 19.2\% of cells, we average the passes before any other aggregation.

\paragraph{Coverage.}
The release annotates 2,828 of our 3,277 documents for at least one system, and our own references on 2,292 of those, giving 7,666 scored segments. All sixteen
pairs are covered. Coverage is complete at 198 documents for \enghye{}, \engjpn{}, \engukr{}, \engtha{} and \engind{}, and thin for \engzhs{} and \engzht{} at 25 and 28 documents. Three hundred and eighty-three annotators
contribute; the largest supplies 3.8\% of annotations.

\paragraph{The unit of aggregation changes the answer.}
The release does not state how the official score aggregates, and the choice is not innocuous for us. Averaging over segments gives our references 75.8 ESA and
thirteenth place of 44 systems; averaging within documents first gives 67.4 and twentieth. Table~\ref{tab:units} shows why. Every spoken document is presented to annotators as a single item, whereas a software document contributes up to 37. The spoken domain is therefore 19.8\% of segments but 66.1\% of documents, and it is the domain where our references are weakest. The shift is the largest of any system in the task: human post-editing moves the other way, from eighth to second. We report both conventions throughout and draw no conclusion that depends on the choice.

\begin{table}[h]
  \centering
  \resizebox{\columnwidth}{!}{%
  \begin{tabular}{@{}lrrrrr@{}}
    \toprule
    Domain & Docs & Segs & Seg/doc & seg ESA & doc ESA \\
    \midrule
    software & 72 & 1{,}825 & 25.4 & 86.8 & 86.3 \\
    news & 179 & 2{,}130 & 11.9 & 82.6 & 82.2 \\
    factchecking & 6 & 32 & 5.3 & 85.2 & 85.5 \\
    social & 509 & 2{,}130 & 4.2 & 68.5 & 71.9 \\
    edu & 10 & 33 & 3.3 & 81.8 & 77.9 \\
    speech & 1{,}516 & 1{,}516 & 1.0 & 63.2 & 63.2 \\
    \midrule
    all & 2{,}292 & 7{,}666 & 3.3 & 75.8 & 67.4 \\
    \bottomrule
  \end{tabular}}
  \caption{How many annotation items each document contributes, by domain, and
  the mean human ESA of our references under the two aggregation conventions.
  The spoken domain is one item per document and drives the difference between
  the two final rows.}
  \label{tab:units}
\end{table}

\paragraph{Position in the evaluation.}
Segment-averaged, our references place thirteenth of 44 at 75.8 [75.2,~76.4]; a bootstrap over segments puts the rank between eleventh and fifteenth in 95\% of resamples, so the position is a band rather than a point. Document-averaged they place twentieth at 67.4 [66.4,~68.5]. Human translation from scratch leads on both conventions. Human post-editing places eighth segment-averaged, below six participating systems, and second document-averaged, which is itself evidence that a single human reference is not a fixed ceiling.

\paragraph{Distance from human translation.}
Table~\ref{tab:humaneval} compares our references against the two human systems on identical segments and identical documents. Outside the spoken domain the segment-averaged gap to human post-editing is 2.2 ESA and the interval excludes zero; document-averaged the same contrast is 1.0 ESA and the interval does not. Two domain cells are indistinguishable from a human translator on paired
segments: software against translation from scratch at $+$0.3 [$-$2.6,~$+$3.0] over 295 segments, and news against post-editing at $+$0.3 [$-$2.0,~$+$2.6] over 437 segments. Elsewhere the difference is in the failure tail, with 15.0\% of our segments below 50 against 10.8\% for human post-editing. Human systems appear in six of our sixteen pairs.

\begin{table}[h]
  \centering
  \resizebox{\columnwidth}{!}{%
  \begin{tabular}{@{}llrrrr@{}}
    \toprule
    Unit & Comparison & n & ours & human & gap (95\% CI) \\
    \midrule
    \multicolumn{6}{@{}l}{\emph{vs.\ Human (from scratch)}} \\
    seg & all & 1{,}363 & 78.0 & 83.7 & $-$5.7 [$-$7.0, $-$4.5] \\
    seg & excl.\ speech & 1{,}096 & 80.2 & 83.8 & $-$3.6 [$-$5.0, $-$2.3] \\
    seg & speech only & 267 & 69.2 & 83.4 & $-$14.3 [$-$17.5, $-$11.0] \\
    doc & all & 414 & 71.9 & 83.2 & $-$11.3 [$-$13.8, $-$9.0] \\
    doc & excl.\ speech & 147 & 77.0 & 82.8 & $-$5.9 [$-$8.6, $-$3.2] \\
    \midrule
    \multicolumn{6}{@{}l}{\emph{vs.\ Human (post-editing)}} \\
    seg & all & 1{,}470 & 74.7 & 80.5 & $-$5.8 [$-$7.2, $-$4.4] \\
    seg & excl.\ speech & 1{,}161 & 77.5 & 79.7 & $-$2.2 [$-$3.7, $-$0.7] \\
    seg & speech only & 309 & 63.9 & 83.4 & $-$19.5 [$-$22.9, $-$16.1] \\
    doc & all & 487 & 69.1 & 81.8 & $-$12.7 [$-$15.4, $-$10.1] \\
    doc & excl.\ speech & 178 & 78.1 & 79.1 & $-$1.0 [$-$4.1, $+$1.9] \\
    \bottomrule
  \end{tabular}}
  \caption{Our references against the human systems, paired on identical
  segments (seg) or identical documents (doc), mean ESA. Intervals are
  percentile bootstrap, 2{,}000 resamples. The speech-only rows are identical
  across units because every spoken document is one segment. Human systems
  appear in \cesdeu{}, \engdeu{}, \engekk{}, \engisl{}, \engjpn{} and
  \zhsjpn{}.}
  \label{tab:humaneval}
\end{table}

\paragraph{Domain profile.}
Table~\ref{tab:humanevaldomain} gives the profile of our own references.
Documents carrying video or screenshots score 65.0 against 82.5 for text-only documents at document level, 65.4 against 84.3 at segment level.
That gap is not ours alone: Gemini~3.1~Pro scores 74.7 on media documents against 88.6 on text-only ones, so multimodal documents are harder for every system.
Our additional deficit on them is 9.3 ESA against 4.3 on text-only documents, and the next paragraph shows where it comes from.

\begin{table}[h]
  \centering
  \small
  \begin{tabular}{@{}lrrrr@{}}
    \toprule
    Domain & Segs & ESA & Median & \% $<$50 \\
    \midrule
    software & 1{,}825 & 86.8 & 95 & 5.5 \\
    factchecking & 32 & 85.2 & 92 & 6.2 \\
    news & 2{,}130 & 82.6 & 90 & 7.7 \\
    edu & 33 & 81.8 & 90 & 6.1 \\
    social & 2{,}130 & 68.5 & 80 & 20.2 \\
    speech & 1{,}516 & 63.2 & 70 & 29.7 \\
    \midrule
    all & 7{,}666 & 75.8 & 85 & 15.0 \\
    \bottomrule
  \end{tabular}
  \caption{Human ESA of our references by domain, segment-averaged. The
  document-averaged column is given in Table~\ref{tab:units}.}
  \label{tab:humanevaldomain}
\end{table}

\paragraph{What the selection step cost.}
Grouping our references by the model family that produced the selected candidate splits them into two populations, and comparing each against the participating systems on the same segments shows that the split is not a property of the documents. Table~\ref{tab:selcostfull} gives the paired gaps. On the 4,912 segments where the selector kept a GPT-5.5 or Gemini candidate our reference is within 2.1 ESA of the six strongest systems and ahead of two of
them. On the 2,541 segments where it kept a candidate from TranslateGemma, HY-MT, Tower-Plus, Aya or EuroLLM it is 14.4 to 17.3 points behind all six. The
anchors themselves barely move between the two groups, 82.8 against 82.5 for Gemini~3.1~Pro, so the open-selected documents were no harder to translate. On spoken documents the contrast is sharper: where a frontier candidate was kept our reference is statistically indistinguishable from five of the six systems and ahead of the sixth, and where an open candidate was kept it is 23.5 to 25.5 points behind all of them. Taking our offset against Gemini~3.1~Pro on the frontier-selected segments and applying it to the anchor level on the open-selected ones estimates 81.1 ESA had a frontier candidate been kept
throughout, giving 81.3 overall against the 75.9 obtained on the same 7,453 segments. That estimate assumes the offset transfers, and is quoted as an order of magnitude rather than a prediction. Producer attribution is by exact text match against the generation outputs and is unambiguous at family level for all but one document; 21 of the 2,292 documents match no cell and are excluded from this analysis.

\begin{table}[h]
  \centering
  \resizebox{\columnwidth}{!}{%
  \begin{tabular}{@{}lrrrrrrr@{}}
    \toprule
    & & \multicolumn{6}{c}{gap against participating system} \\
    \cmidrule(l){3-8}
    Group & Ours & Gem. & TRIVE & Wayf. & Lumen & GPT & UvA \\
    \midrule
    \multicolumn{8}{@{}l}{\emph{all domains}} \\
    frontier, 4{,}912 & 81.4 & $-$1.4 & $-$1.7 & $-$1.9 & $-$2.1 & $+$0.9 & $+$0.1 \\
    open MT, 2{,}541 & 65.2 & $-$17.3 & $-$16.5 & $-$16.1 & $-$15.2 & $-$14.4 & $-$14.4 \\
    \midrule
    \multicolumn{8}{@{}l}{\emph{spoken documents}} \\
    frontier, 584 & 80.1 & $+$0.4 & $+$0.5 & $+$0.9 & $+$3.5 & $+$1.7 & $+$1.1 \\
    open MT, 931 & 52.5 & $-$25.5 & $-$24.1 & $-$24.0 & $-$23.7 & $-$23.7 & $-$23.5 \\
    \bottomrule
  \end{tabular}}
  \caption{Mean human ESA of our references by the family of the selected
  candidate, and the paired gap against each participating system on identical
  segments. In the frontier rows every gap except Lumen on spoken documents has
  a 95\% interval containing zero or close to it; in the open rows every
  interval excludes zero by a wide margin. Widths are given in the released
  verification output.}
  \label{tab:selcostfull}
\end{table}

\paragraph{Which language pairs were hardest.}
Ranking pairs by raw ESA is misleading, because the pairs differ in domain mix
and in how much of each was annotated. \engdeu{} has our lowest raw score at
59.7 but is only mid-table once the comparison is controlled, while \enghye{}
has both the highest raw score and the smallest deficit. The measure that
answers the question is the paired gap against the six strongest participating
systems on identical segments, given in Table~\ref{tab:humanevalpairs}. Our
references are furthest behind on \engzhs{} ($-11.3$), \engkor{} ($-11.1$),
\engjpn{} ($-9.5$), \engdeu{} ($-9.3$), \engrus{} ($-9.2$) and \engbel{}
($-9.0$), and closest on \enghye{} ($-1.2$, whose interval includes zero, as does the much wider one for \engzht{}), \zhsjpn{} ($-2.7$), \engkaz{} ($-3.1$), \engekk{} ($-3.2$) and
\engisl{} ($-3.4$). The five thinnest pairs, \engzhs{} and \engzht{} at 25 and
28 documents and \engkor{}, \engdeu{} and \cesdeu{} at 51, 68 and 70, carry
intervals two to four times wider than the rest and should not be read as
point estimates.
 
The ordering tracks the share of segments on which the selector kept an
open-weight translation model rather than a frontier one: Pearson $-0.56$ and
Spearman $-0.65$ across the sixteen pairs. It is a tendency and not a rule.
\engbel{} is the informative exception, at only 17\% open-MT selection yet
$-9.0$ overall, because both slices are weak there: its 75 open-MT segments
score 41.5 against an anchor level of 65.5, and even its 357 frontier segments
sit 5.9 below the anchors, against 1.4 globally. Belarusian is hard for us
whatever we select.
 
The three pairs that motivated the language penalty behave in a way the
penalty explains. On \enghye{} and \engkaz{} the penalty pushed selection
almost entirely onto frontier models, 3\% and 16\% open-MT respectively, and
those are two of our three strongest pairs; on \enghye{} our references sit
0.4 below the anchors on the frontier-selected segments, which is
indistinguishable from the best systems in the task. Removing wrong-language
output and improving quality turn out to be the same intervention here, which
was not something the construction-time evidence could have shown.
 
Document averaging changes the magnitudes substantially, \engkor{} moving from
$-11.1$ to $-26.5$, but preserves most of the ordering, with Spearman $+0.81$
between the two columns.
 
\begin{table}[h]
  \centering
  \resizebox{\columnwidth}{!}{%
  \begin{tabular}{@{}lrrrrrr@{}}
    \toprule
    Pair & Docs & Segs & ESA & \% open & gap (95\% CI) & doc gap \\
    \midrule
    \engzhs{} & 25 & 70 & 66.0 & 66 & $-$11.3 [$-$16.7, $-$5.9] & $-$12.8 \\
    \engkor{} & 51 & 167 & 69.9 & 27 & $-$11.1 [$-$15.3, $-$7.0] & $-$26.5 \\
    \engjpn{} & 198 & 691 & 79.0 & 55 & $-$9.5 [$-$11.3, $-$8.0] & $-$14.2 \\
    \engdeu{} & 68 & 173 & 59.7 & 61 & $-$9.3 [$-$12.6, $-$6.0] & $-$13.5 \\
    \engrus{} & 189 & 664 & 72.4 & 48 & $-$9.2 [$-$11.2, $-$7.5] & $-$17.5 \\
    \engbel{} & 136 & 435 & 67.5 & 17 & $-$9.0 [$-$11.4, $-$6.8] & $-$13.5 \\
    \engind{} & 198 & 691 & 74.8 & 54 & $-$7.8 [$-$9.4, $-$6.1] & $-$13.9 \\
    \engtha{} & 198 & 691 & 77.0 & 26 & $-$7.3 [$-$8.8, $-$5.7] & $-$17.5 \\
    \cesdeu{} & 70 & 192 & 78.4 & 53 & $-$6.2 [$-$9.5, $-$3.0] & $-$7.5 \\
    \engukr{} & 198 & 691 & 78.9 & 44 & $-$5.7 [$-$7.4, $-$4.2] & $-$13.4 \\
    \engzht{} & 28 & 88 & 65.3 & 49 & $-$3.9 [$-$9.7, $+$1.8] & $-$8.6 \\
    \engisl{} & 192 & 647 & 79.1 & 24 & $-$3.4 [$-$4.9, $-$1.9] & $-$7.2 \\
    \engekk{} & 172 & 591 & 77.1 & 24 & $-$3.2 [$-$4.6, $-$1.7] & $-$8.3 \\
    \engkaz{} & 181 & 643 & 78.4 & 16 & $-$3.1 [$-$4.6, $-$1.7] & $-$4.3 \\
    \zhsjpn{} & 190 & 541 & 72.9 & 36 & $-$2.7 [$-$4.4, $-$1.2] & $-$4.0 \\
    \enghye{} & 198 & 691 & 81.0 & 3 & $-$1.2 [$-$2.6, $+$0.3] & $-$3.0 \\
    \midrule
    all & 2{,}292 & 7{,}666 & 75.8 & 34 & $-$5.9 [$-$6.4, $-$5.4] & $-$11.4 \\
    \bottomrule
  \end{tabular}}
  \caption{Per-pair difficulty, ordered by the segment-level gap. ESA is our
  own segment-averaged score; \% open is the share of annotated segments on
  which the selector kept a translation-specialised open model; gap is our ESA
  minus the mean of Gemini~3.1~Pro, TRIVE, Wayfinder, Lumen, GPT-5.5 and
  UvA-MT on identical segments, with a percentile bootstrap interval over
  2{,}000 resamples; doc gap is the same quantity computed on document means.
  Segments are counted where at least four of the six anchors were annotated,
  which is 7,644 of our 7,666.}
  \label{tab:humanevalpairs}
\end{table}

\paragraph{The same result without conditioning on our selector.}
Grouping by what we selected invites the objection that the grouping is itself a consequence of the metric. A test that avoids it: for each annotated document, compare the best frontier candidate with the best open candidate under MetricX and group by which the metric preferred. MetricX preferred an open candidate on 1,579 of the 2,292 documents. On those 4,170 segments, our references score 69.8 [68.9,~70.6] while Gemini~3.1~Pro scores 80.8, TRIVE 80.2, Wayfinder 80.0 and GPT-5.5 78.2. On the 713 documents where MetricX preferred a frontier candidate, our references score 83.1 [82.4,~83.8] against 84.3, 85.0, 85.5 and 82.9. Where the metric preferred an open candidate, we ended around ten points below the achievable level; where it preferred a frontier candidate, we were within two. Note that C6 is not a MetricX argmax, so on the first group we still shipped a frontier candidate for 1,651 of 4,032 segments.

\paragraph{Which cells the metric misranks.}
The effect is not a general preference for open-weight models. Computed over
all candidates on the same documents rather than only the selected ones
(Table~\ref{tab:famrank}), MetricX penalises Tower-Plus, Aya and EuroLLM heavily and correctly. What it does is rate TranslateGemma best of all seven families, ahead of both frontier models, and HY-MT close behind them. Those two cells produced 941 of our annotated documents at 54.1 and 52.8 human ESA against 80.9 and 79.0 for the frontier families. Across the seven families, the rank correlation between human ESA and MetricX is $+0.07$, that is, absent rather than inverted; the strongly negative value obtained when the metric is averaged only over the documents where each family won selection is an artefact of that conditioning, since a family wins precisely when its score is unusually good.

\begin{table}[h]
  \centering
  \small
  \begin{tabular}{@{}lrrrr@{}}
    \toprule
    Family & Docs & ESA & MX all & MX sel. \\
    \midrule
    TranslateGemma & 564 & 54.1 & 4.63 & 3.29 \\
    GPT-5.5 & 578 & 79.0 & 5.06 & 5.12 \\
    Gemini-3.1-Pro & 506 & 80.9 & 5.11 & 4.98 \\
    HY-MT1.5 & 377 & 52.8 & 5.40 & 2.60 \\
    Tower-Plus & 122 & 63.9 & 9.51 & 3.26 \\
    Aya-Expanse & 89 & 67.1 & 12.53 & 4.56 \\
    EuroLLM & 35 & 62.6 & 13.03 & 3.53 \\
    \bottomrule
  \end{tabular}
  \caption{Human ESA of our references by producing family, document-averaged, against mean MetricX (lower is better) computed two ways: over every candidate that family produced for these documents (MX all), and over only the documents where that family won selection (MX sel.). The second column must not be used to rank families.}
  \label{tab:famrank}
\end{table}

\paragraph{Our selection metrics did not predict the human scores.}
Scoring our own selected references, the document-level correlation between our
QE metrics and human ESA is near zero once multimodality is controlled:
MetricX reaches $+0.07$ on text-only documents and CometKiwi $+0.19$, CometKiwi
reaches $+0.19$ within news, and MetricX $+0.27$ within the social domain. The
uncontrolled correlation is slightly negative, $-0.12$ for MetricX over 2,272
documents. This agrees with Section~\ref{sec:calibration}: among top candidates
these metrics carry little information about which one a human will prefer.

\paragraph{Error spans.}
Averaging annotation passes within a cell, our references carry 1.58 spans per
segment and 0.55 major spans, against 1.22 and 0.30 for human translation from
scratch and 0.81 and 0.36 for human post-editing. Span rate on our references
tracks the domain profile: 3.16 per segment on spoken documents, 1.85 on
social, 1.05 on news and 0.57 on software. Span lengths are not directly
comparable across systems, because 12.7\% of all spans are zero-width
insertion markers and the rate varies from 7.9\% for human translation from
scratch to 19.7\% for human post-editing.

\paragraph{Checks on the comparison.}
Systems are annotated on overlapping but different subsets, so every comparison
above is restricted to segments or documents where both were scored, and the
leaderboard position is descriptive only. Annotator identity is not a
confound: where the same annotator scored both our reference and the human one,
the gap is close to the gap across annotators ($-6.4$ against $-5.3$ for
translation from scratch, $-5.1$ against $-6.2$ for post-editing). The human
systems are distinct text, sharing no document rendering with our references on
any of the 414 and 487 documents where both are annotated. The successive
releases of 12 August, 19 August and 2 September are cumulative at the level of
annotator, item and system coverage, so the analysis was regenerated rather
than merged; we have not verified that scores attached to retained items are
unchanged between releases.

\end{document}